\documentclass[11pt]{article}

\usepackage{booktabs} 

\usepackage[a4paper]{geometry}

\usepackage[utf8]{inputenc}
\usepackage{makeidx}
\usepackage[T1]{fontenc}
\usepackage[english]{babel}
\usepackage{graphicx}
\usepackage{amsfonts,amsmath,amssymb,amsthm}
\usepackage{lmodern}

\usepackage{mathrsfs}
\usepackage[active]{srcltx}
\usepackage{verbatim}
\usepackage{enumitem} 
\usepackage[toc,page]{appendix}
\usepackage{aliascnt,bbm}
\usepackage{array}
\usepackage{xargs}
\usepackage{cellspace}
\usepackage{xcolor}

\usepackage{algorithm}
\usepackage{algorithmic}

\usepackage[Symbolsmallscale]{upgreek}
\usepackage{xstring}

\providecommand{\keywords}[1]
{
  \small	
  \textbf{\textit{Keywords---}} #1
}

\usepackage{accents}
\usepackage{dsfont}
\usepackage{aliascnt}
\usepackage{cleveref}
\makeatletter

\crefname{theorem}{theorem}{Theorems}
\Crefname{Theorem}{Theorem}{Theorems}

\newaliascnt{lemma}{theorem}

\aliascntresetthe{lemma}
\crefname{lemma}{lemma}{lemmas}
\Crefname{Lemma}{Lemma}{Lemmas}

\newaliascnt{corollary}{theorem}

\aliascntresetthe{corollary}
\crefname{corollary}{corollary}{corollaries}
\Crefname{Corollary}{Corollary}{Corollaries}

\newaliascnt{proposition}{theorem}

\aliascntresetthe{proposition}
\crefname{proposition}{proposition}{propositions}
\Crefname{Proposition}{Proposition}{Propositions}

\newaliascnt{definition}{theorem}

\aliascntresetthe{definition}
\crefname{definition}{definition}{definitions}
\Crefname{Definition}{Definition}{Definitions}

\newaliascnt{remark}{theorem}

\aliascntresetthe{remark}
\crefname{remark}{remark}{remarks}
\Crefname{Remark}{Remark}{Remarks}

\crefname{example}{example}{examples}
\Crefname{Example}{Example}{Examples}

\crefname{figure}{figure}{figures}
\Crefname{Figure}{Figure}{Figures}

\Crefname{assumption}{\textbf{H}\hspace{-3pt}}{\textbf{H}\hspace{-3pt}}
\crefname{assumption}{\textbf{H}}{\textbf{H}}

\makeatother

\def\dataset{\mathcal{D}}
\def\datatest{\mathcal{T}}
\def\datarepr{\mathcal{R}}
\def\databoot{\mathcal{R}_B}
\def\spacex{\mathcal{X}}
\def\spacey{\mathcal{Y}}
\def\spacez{\mathcal{Z}}
\def\sx{x}
\def\sz{z}
\def\sX{X}
\def\sy{y}
\def\sY{Y}
\def\nZ{Z}
\newcommandx{\classifier}[1][1=]{f_{#1}}
\newcommand{\classifierhat}{\hat{f}}
\newcommandx{\pclassifier}[1][1=]{g_{#1}}

\def\param{\theta}

\def\spaceparam{\Theta}
\def\nbclass{K}
\def\ndata{N}
\def\ntest{n_{\operatorname{test}}}
\def\nsamples{n_{\operatorname{samples}}}
\newcommand{\pb}[2]{p\left(#1 \middle| #2 \right)}
\newcommand{\prior}[1]{p\left(#1 \right)}
\newcommand{\pbq}[2]{q\left(#1 \middle| #2 \right)}
\newcommand{\pbhat}[2]{\hat{p}\left(#1 \middle| #2 \right)}
\newcommand{\pbqhat}[2]{\hat{q}\left(#1 \middle| #2 \right)}

\def\Ccal{\mathcal{C}}
\def\Acal{\mathcal{A}}
\def\ECE{\operatorname{ECE}}
\def\MCE{\operatorname{MCE}}
\def\retain{r}

\def\cov{\operatorname{cov}}

\def\srisk{\operatorname{srisk}}
\def\conf{\kappa}
\def\thres{s}
\def\datathres{\mathcal{S}}

\def\AURC{\operatorname{AURC}}

\def\TP{\operatorname{TP}}
\def\TN{\operatorname{TN}}
\def\FP{\operatorname{FP}}
\def\FN{\operatorname{FN}}
\def\TPR{\operatorname{TPR}}
\def\FPR{\operatorname{FPR}}

\def\softmax{\operatorname{softmax}}
\def\weights{W}
\def\bias{b}

\def\SGLD{\operatorname{SGLD}}
\def\SGD{\operatorname{SGD}}
\def\paramstar{\param^*}
\def\batch{\mathcal{S}}
\def\bsize{s}
\def\lr{\gamma}
\def\pdrop{p_{\operatorname{drop}}}
\def\nepochs{n_{\operatorname{epochs}}}
\def\nthinning{n_{\operatorname{thinning}}}

\def\SR{\operatorname{SR}}
\def\SRhat{\widehat{\operatorname{SR}}}
\def\STD{\operatorname{STD}}
\def\STDhat{\widehat{\operatorname{STD}}}

\def\neglogl{\operatorname{neglogl}}

\def\card{\operatorname{card}}

\newcommand{\argmax}{\operatorname*{arg\,max}}

\newcommandx{\VarDeux}[3][3=]{\operatorname{Var}^{#3}_{#1}\left[#2 \right]}

\newcommand{\PE}{\mathbb{E}}
\newcommand{\PP}{\mathbb{P}}

\newcommand{\absolute}[1]{\left\vert #1 \right\vert}

\newcommandx{\Vnorm}[2][1=V]{\| #2 \|_{#1}}
\newcommandx{\VnormEq}[2][1=V]{\left\| #2 \right\|_{#1}}
\newcommandx{\norm}[2][1=]{\ifthenelse{\equal{#1}{}}{\left\Vert #2 \right\Vert}{\left\Vert #2 \right\Vert^{#1}}}

\newcommandx{\normLigne}[2][1=]{\ifthenelse{\equal{#1}{}}{\Vert #2 \Vert}{\Vert #2\Vert^{#1}}}

\newcommand{\parenthese}[1]{\left(#1 \right)}

\newcommand{\defEns}[1]{\left\lbrace #1 \right\rbrace }

\newcommandx\probaMarkovTilde[2][2=]
{\ifthenelse{\equal{#2}{}}{{\widetilde{\mathbb{P}}_{#1}}}{\widetilde{\mathbb{P}}_{#1}\left[ #2\right]}}

\newcommand{\expe}[1]{\PE \left[ #1 \right]}

\newcommand{\probacond}[2]{\PP \left( #1 \middle| #2 \right)}

\def\ie{\textit{i.e.}}

\def\eqsp{\;}

\newcommand{\coint}[1]{\left[#1\right)}

\newcommand{\ooint}[1]{\left(#1\right)}
\newcommand{\ccint}[1]{\left[#1\right]}

\def\rmd{\mathrm{d}}
\newcommand{\wrt}{w.r.t.}

\def\iid{i.i.d.}

\def\eg{e.g.}

\def\rset{\mathbb{R}}

\def\nset{\mathbb{N}}

\newcommandx{\CPE}[3][1=]{{\mathbb E}^{#3}_{#1}\left[#2 \right]} 
\newcommandx{\CPVar}[3][1=]{\mathrm{Var}^{#3}_{#1}\left\{ #2 \right\}}
\newcommand{\CPP}[3][]
{\ifthenelse{\equal{#1}{}}{{\mathbb P}\left(\left. #2 \, \right| #3 \right)}{{\mathbb P}_{#1}\left(\left. #2 \, \right | #3 \right)}}

\newcommand{\1}{\mathbbm{1}}

\newcommandx{\wasser}[1][1=2]{\operatorname{W}_{#1}}

\title{On Last-Layer Algorithms for Classification: \\ Decoupling Representation from Uncertainty Estimation}

\author{Nicolas Brosse\textsuperscript{1}, Carlos Riquelme\textsuperscript{2}, Alice Martin\textsuperscript{1}, Sylvain Gelly\textsuperscript{2} , \'Eric Moulines\textsuperscript{1}}
\date{\today}

\begin{document}

\footnotetext[1]{Centre de Math\'ematiques Appliqu\'ees, UMR 7641, Ecole Polytechnique, France. \\
Emails: nicolas.brosse@polytechnique.edu, alice.martin@polytechnique.edu, \\ eric.moulines@polytechnique.edu. \\
This work was done while Nicolas was an intern at Google Brain.}
\footnotetext[2]{Google Brain Zürich
Emails: rikel@google.com, sylvaingelly@google.com}

\maketitle

\begin{abstract}
\noindent Uncertainty quantification for deep learning is a challenging open
problem. Bayesian statistics offer a mathematically grounded framework to reason
about uncertainties; however, approximate posteriors for modern neural networks
still require prohibitive computational costs. We propose a family of algorithms
which split the classification task into two stages: representation learning and
uncertainty estimation. We compare four specific instances, where uncertainty
estimation is performed via either an ensemble of Stochastic Gradient Descent or
Stochastic Gradient Langevin Dynamics snapshots, an ensemble of bootstrapped
logistic regressions, or via a number of Monte Carlo Dropout passes. We evaluate
their performance in terms of \emph{selective} classification (risk-coverage),
and their ability to detect out-of-distribution samples. Our experiments suggest
there is limited value in adding multiple uncertainty layers to deep
classifiers, and we observe that these simple methods strongly outperform a
vanilla point-estimate SGD in some complex benchmarks like ImageNet.
\end{abstract}


\section{Introduction}
\label{sec:introduction}


The most popular application of deep learning involves the use of a single model trained to convergence by some stochastic optimization method on a supervised dataset. It is hard to deny that this approach has led to impressive wins in a variety of industries. The reason deep models are successful is mainly related to their predictive power, their predictions are \emph{usually} right, \ie\, the models are accurate. The latter is an average statement, and, unfortunately, at the individual data-point level, it is often difficult to know what the \emph{confidence} of the model in its own prediction is. Accordingly, deep systems are currently being deployed in scenarios where making mistakes is cheap. However, before machine learning widens its adoption to fields with critical use-cases, we need to develop systems that are able to say ``I don't know'' when their prediction is likely to be wrong.

More concretely, deep models are now applied to diverse fields such as physics \cite{pmlr-v42-cowa14,Radovic2018MachineLA,he2018analysis}, biology \cite{anjos:iglesias:2015}, healthcare \cite{leibig2017leveraging,Nair2018ExploringUM,liu2018artificial}, or autonomous driving \cite{KendallGal2017UncertaintiesB,2018arXiv181106817M} to name a few.
In these cases, quantifying and processing model uncertainty is of crucial importance \cite{Krzywinski2013,2016arXiv160606565A}, as the main goal is to automate decision making while providing strong risk guarantees. Properly calibrated confidence functions should enable the identification of inputs for which predictions are likely to be erroneous, and that should be for instance flagged for human intervention.

\begin{figure}
\begin{center}
\includegraphics[scale=0.5]{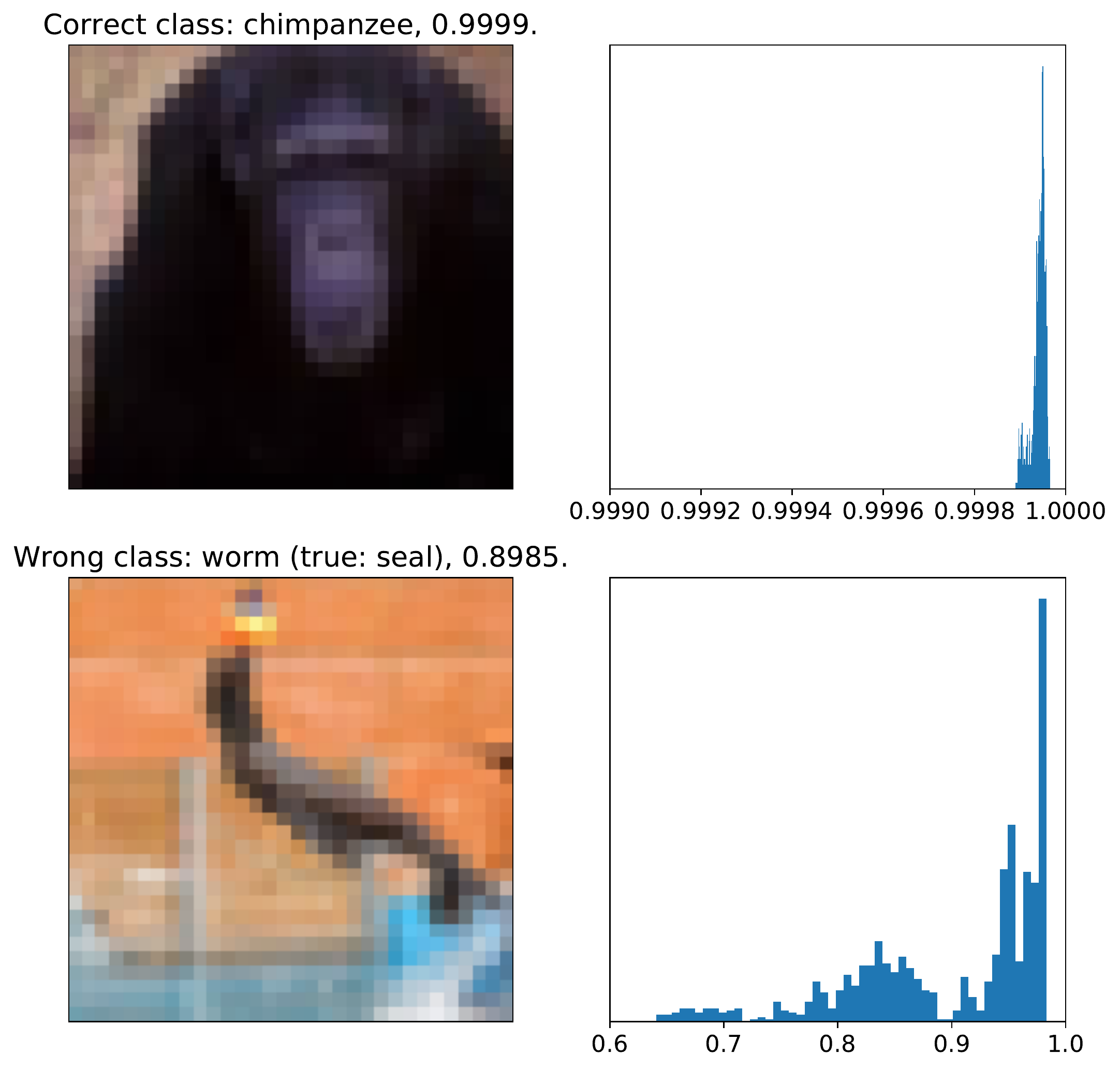}
\caption{Histograms of $p_{\max}$ values given by Stochastic Gradient Langevin Dynamics (SGLD) on top of a pre-trained VGG-16 network on CIFAR-100, $\defEns{\max_k \pb{k}{\sx, \param_i}}_{i=1}^{\nsamples}$. See $x-$axis.
\textbf{Top row:} Chimpanzee image, correctly classified.
The values are extremely concentrated around the average $\bar{p}_{\max} = 0.9999$.
\textbf{Bottom row:} Seal image, wrongly classified as a worm.
Class is predicted with a high average softmax ($\bar{p}_{\max} = 0.8985$), but the histogram shows a larger standard deviation of predictions, a valuable estimate of uncertainty.}
\label{figure:hist-cifar-100}
\end{center}
\end{figure}

The softmax probabilities outputted by deep classifiers can be erroneously interpreted as prediction confidence. Unfortunately, sometimes, high confidence predictions can be woefully incorrect, and fail to indicate when they are likely mistaken; see \cite{43405} and the references therein.
\Cref{figure:hist-cifar-100} shows an example of this; a seal picture is wrongly classified as a worm, whereas its softmax value is $p_{\max} = 0.90$.

As a consequence, uncertainty quantification for deep learning is an active area of research.
The Bayesian framework offers a principled approach to do probabilistic inference \cite{Hinton:1993:KNN:168304.168306,Neal:1996:BLN:525544,
ensemble-learning-in-bayesian-neural-networks}; however, at the scale of modern deep neural networks, even approximate Bayesian and frequentist methods face serious computational issues \cite{Gal:2016:DBA:3045390.3045502,NIPS2017_7219}.


In this paper, we propose a family of simple classification algorithms that provide uncertainty quantification at a modest additional computational cost.
The basic idea is as follows: after training end-to-end a deep classifier on input-output pairs ($\sx, \sy$) to obtain an accurate task-dependent representation $\sz$ of the data, we then fit an ensemble of models on $(\sz, \sy)$.
The simplicity of this new dataset allows us to compute explicit uncertainty estimates.
In particular, we explore four concrete instances of uncertainty algorithms, based on Stochastic Gradient Descent \cite{Mandt:2017:SGD:3122009.3208015}, Stochastic Gradient Langevin Dynamics \cite{Welling:2011:BLV:3104482.3104568}, the Bootstrap, see Section 8.4 of \cite{friedman2001elements}, and Monte Carlo Dropout \cite{Gal:2016:DBA:3045390.3045502}.
The core idea has some connections with transfer learning \cite{Yosinski:2014:TFD:2969033.2969197,Razavian:2014:CFO:2679599.2679731,pmlr-v32-donahue14}. By sequentially tackling two tasks (representation learning and uncertainty quantification), these algorithms performed on the last layer of the neural networks reduce the computational cost associated with the approximated inference compared to their full network versions.

Our experiments suggest that there is limited value in adding multiple uncertainty layers to high-level representations in deep classifiers.\footnote{The code, implemented in Keras \cite{chollet2015keras} and Tensorflow \cite{tensorflow2015-whitepaper}, is available at \url{https://github.com/nbrosse/uncertainties}.}
As expected, in terms of selective classification, last-layer algorithms outperform a point-estimate network baseline trained on SGD in datasets like ImageNet.

\section{Related Work}
\label{sec:related-work}

Uncertainty estimation has a rich history, and we describe here the work most
closely related to ours. A review of Bayesian neural networks is provided in
\cite{gal:2016}. In particular, \cite{Gal:2016:DBA:3045390.3045502} proposes
Monte-Carlo dropout, a Bayesian technique for estimating uncertainties in neural
networks by applying dropout at test time.

Frequentist approaches mainly focus on \emph{selective classification},
\emph{calibration}, and \emph{out-of-distribution} detection. These concepts are
introduced and detailed in the following sections. In particular, selective
classification and out-of-distribution rely on a confidence function which
outputs a score of confidence, in addition to the predicted class. In selective
classification, uncertain inputs are considered as rejected or left out by the
classifier, which enables to draw a risk-coverage curve. In out-of-distribution,
uncertain inputs are considered as out-of-distribution samples. A simple way to
deal with selective classification is by using the softmax value of the chosen
class as the confidence function. It has been shown to outperform MC-Dropout on
ImageNet \cite{geifman:yaniv:2017}. A general technique on top of an uncertainty
estimate is developed in \cite{geifman:uziel:yaniv:2018}, and a novel loss is
introduced in \cite{thulasidasan2019knows} to train neural networks to abstain
from predicting. A direct optimization of the ROC curve (for the binary decision
abstention/classification) is presented in \cite{2018arXiv180207024A}. We
address calibration in \Cref{sec:metrics} and out-of-distribution detection in
\Cref{sec:suppl-out-of-distribution}.



The \emph{ensemble} technique that we adopt in this paper has shown satisfactory results both for classical metrics (such as accuracy) and uncertainty-related ones \cite{2015arXiv151106314L,NIPS2016_6205,NIPS2017_7219,2018arXiv180210026G,2017arXiv170400109H}.

In the context of decision making, the idea of using the last layer of a pre-trained regression neural network to compute uncertainty estimates has been explored in several related fields: Bayesian optimization \cite{snoek2015scalable}, active learning \cite{2018arXiv181112535Z}, and as an uncertainty source for exploration in reinforcement learning \cite{riquelme2018deep, azizzadenesheli2018efficient}.
Combining neural networks with Gaussian processes has also been suggested as a way to decouple representation and uncertainty \cite{calandra2016manifold,2017arXiv170705922I}.

\section{Problem Description}
\label{sec:setup}

In this work, we study classification tasks. Let $\spacex$ be a feature space, and $\spacey = \defEns{1,\ldots, \nbclass}$ a  finite label set with $\nbclass \geq 2$ classes.
We assume access to a training dataset $\dataset = \defEns{\sx_i,\sy_i}_{i=1}^{\ndata} \subset \parenthese{\spacex, \spacey}^{\ndata}$ of $\ndata$ points independently distributed according to a pair of random variables $(\sX, \sY)$. We define the test set analogously, $\datatest = \defEns{\sx_i, \sy_i}_{i=1}^{\ntest}$.
For classification tasks, the standard output of a neural network provides a probability distribution over the $\nbclass$ classes, by applying the softmax function to the final logits.
Let us denote by $\param$ the set of real-valued parameters of the network (weights and bias).
The network is usually trained using variants of stochastic gradient descent with the cross-entropy loss (the negative log likelihood of the multinomial logistic regression model):
$\neglogl_{\ndata}(\param) = - \frac{1}{\ndata} \sum_{i=1}^{\ndata} \sum_{k=1}^{\nbclass} \1\defEns{\sy_i=k} \log(\pb{k}{\sx_i, \param})$
where $\defEns{\pb{k}{\sx_i, \param}}_{k=1}^{\nbclass}$ is the output probability distribution over $\spacey$ predicted by the network.

The classifier $\classifier[\param]:\spacex\to\spacey$ is generally obtained just by taking the argmax, $\classifier[\param](\sx) = \argmax_{k\in\defEns{1,\ldots,\nbclass}} \pb{k}{\sx, \param}$ for $\sx\in\spacex$. This rule corresponds to the optimal decision when the misclassification cost is independent of the classes, while it can be easily generalized to heterogeneous costs, see \eg~Section 1.5.1 of \cite{bishop:2006}. The performance of $\classifier[\param]$ can be measured by the accuracy; however, to take advantage of uncertainty estimates associated to the classifier $\classifier[\param]$, other metrics need to be defined.

\subsection{Uncertainty Metrics}
\label{sec:metrics}
Neural networks for classification tasks output a probability distribution over $\spacey$.
The notion of \emph{calibration} is thus relevant: a model is calibrated if, on average over input points $\sx\in\spacex$, the predicted distribution $\defEns{\pb{k}{\sx, \param}}_{k=1}^{\nbclass}$ does match the true underlying distribution over the $\nbclass$ classes (note that, in most works, the authors focus on $p_{\max}$ matching only).
When calibrated, the output provides an appropriate measure of uncertainty associated to the decision $\classifier[\param](\sx)$.
However, despite strong accuracies, modern neural networks are often miscalibrated.
Fortunately, remarkably simple methods exist to alleviate this issue, such as temperature scaling, \cite{guo:pleiss:sun:weinberger:2017}. Calibrated neural networks are important for model interpretability; however, they do not offer a systematic and automated way to neither improve accuracy nor detect out-of-distribution samples.

Selective classification is a key metric to measure quality of uncertainty estimates.
It is also sometimes referred to as abstention, and the concept is not restricted to deep learning \cite{Bartlett:2008:CRO:1390681.1442792,NIPS2016_6336,2018arXiv180308355G}.
A selective classifier is a pair $(\classifier, \retain)$ where $\classifier$ is a classifier, and $\retain:\spacex\to\defEns{0,1}$ is a selection function which serves as a binary qualifier for $\classifier$, see \eg~\cite{geifman:yaniv:2017,geifman:uziel:yaniv:2018}. The selective classifier abstains from prediction at a point $x\in\spacex$ if $\retain(x)=0$, and outputs $\classifier(x)$ when $\retain(x)=1$. The performance of a selective classifier can be quantified using the notions of coverage and selective risk.
The coverage is defined as $\cov(\retain) = \expe{\retain(\sX)}$, whereas selective risk is given by
\begin{equation*}
  \srisk(\classifier, \retain) = \frac{\expe{\1\defEns{\sY \neq \classifier(\sX)} \retain(\sX)}}{\expe{\retain(\sX)}} \eqsp.
\end{equation*}
Their empirical estimations over the test set $\datatest$ are:
\begin{align*}
  \cov_{\ntest}(\retain) &= \frac{1}{\ntest}\sum_{i=1}^{\ntest} \retain(\sx_i) \eqsp , \\
  \srisk_{\ntest}(\classifier, \retain) &= \frac{\sum_{i=1}^{\ntest} \1\defEns{\sy_i \neq \classifier(\sx_i)} \retain(\sx_i)}{\sum_{i=1}^{\ntest} \retain(\sx_i)} \eqsp.
\end{align*}
A natural way to define a selection function $\retain$ is by means of a \emph{confidence} function $\conf:\spacex\to\rset$ which quantifies how much we trust the $\classifier(\sx)$ prediction for input $\sx$.
The selection function $\retain$ is then constructed by thresholding $\conf$, \ie~given $\thres\in\rset$, for all $\sx\in\spacex$, we set $\retain_{\thres}(\sx) = \1\defEns{\conf(\sx) \geq \thres}$.
We only classify $\sx$ if its confidence is at least $\thres$.
Let $\datathres$ be the set of all $\conf$ values for those points in the test dataset $\datatest$, $\datathres = \defEns{\conf(\sx), \sx\in\datatest_{\sx}}$, where $\datatest_{\sx}$ is the projection of $\datatest$ over the first coordinate. If there are duplicate values in $\datathres$, they are replicated so that $\card(\datathres) = \ntest$.
The performance of confidence function $\conf$ can be measured using the Area Under the Risk-Coverage curve (AURC) computed over $\datatest$:
\begin{equation*}
  \AURC(\classifier, \conf) = \frac{1}{\ntest} \sum_{\thres\in\datathres} \srisk_{\ntest}(\classifier, \retain_{\thres}) \eqsp.
\end{equation*}
Better confidence functions lead to a faster decrease of the associated risk when we decrease coverage, which results in a lower AURC.
They are able to improve accuracy by choosing not to classify points where uncertainty is highest  and errors are likely.

Concerning \emph{out-of-distribution detection}, we present several standard metrics in \Cref{sec:suppl-out-of-distribution}.

\subsection{Confidence Functions}
\label{sec:confidence}

Selective classification relies on a \emph{confidence function} $\conf$, which quantifies the confidence in the class prediction $\classifier(\sx)$. We present now several ways to define $\conf$; note they are linked to the algorithms we present in \Cref{sec:algorithms}.
First, we introduce some required background concepts.

In the Bayesian framework, a major obstacle often encountered in practice is to sample from the posterior distribution $\param\mapsto\pb{\param}{\dataset}$ where $\param$ denotes the parameters of either full or last layer networks.
Closed-form updates are usually not available, leading to an intractable problem (except for conjugate distributions).
Posteriors can be approximated using workarounds such as variational inference \cite{Wainwright:2008:GME:1498840.1498841}, or Markov Chain Monte Carlo algorithms, see \eg~Chapter 11 of \cite{gelman2013bayesian}. The predictive posterior distribution is defined for $\sx\in\spacex$ and $\sy\in\spacey$ by
\begin{equation}\label{eq:def-predictive-posterior}
  \pb{\sy}{\sx} = \int_{\spaceparam} \pb{\sy}{\sx, \param} \pb{\param}{\dataset} \rmd \param \eqsp,
\end{equation}
where $\sy\mapsto\pb{\sy}{\sx, \param}$ is the likelihood function (the softmax output of the network), and $\spaceparam$ the parameter space. We estimate this quantity in practice by
\begin{equation}\label{eq:def-pbhat}
  \pbhat{\sy}{\sx} = \frac{1}{\nsamples} \sum_{i=1}^{\nsamples} \pb{\sy}{\sx, \param_i} \eqsp,
\end{equation}
where $\defEns{\param_i}_{i=1}^{\nsamples}$ are approximately drawn according to the posterior distribution.
In \Cref{sec:algorithms}, we propose four algorithms from which we can sample $\defEns{\param_i}_{i=1}^{\nsamples}$. The three confidence functions considered are introduced below. 

\paragraph*{Softmax Response.}
The first confidence function we examine is the softmax response (SR) \cite{geifman:yaniv:2017}, also known as (one minus) the variation ratio, p.40-43 of \cite{freeman:1965}.
It is defined for $\sx\in\spacex$ by $\SR(\sx) = \max_{k\in\defEns{1,\ldots,\nbclass}} \pb{k}{\sx}$ where $\defEns{\pb{k}{\sx}}_{k=1}^{\nbclass}$ is the predictive posterior distribution given in \eqref{eq:def-predictive-posterior}. $\SR(\sx)$ is estimated by
\begin{equation}\label{eq:def-SR-empirical}
  \SRhat(\sx) = \max_{k\in\defEns{1,\ldots,\nbclass}} \pbhat{k}{\sx} \eqsp,
\end{equation}
where $\defEns{\pbhat{k}{\sx}}_{k=1}^{\nbclass}$ is defined in \eqref{eq:def-pbhat}.
The associated classifier is then determined by the optimal decision rule $\classifier(\sx) = \argmax_{k\in\defEns{1,\ldots,\nbclass}} \pb{k}{\sx}$.
Its empirical version is
\begin{equation}\label{eq:def-classifier-argmax-empirical}
  \classifierhat(\sx) = \argmax_{k\in\defEns{1,\ldots,\nbclass}} \pbhat{k}{\sx} \eqsp.
\end{equation}

\paragraph*{Standard deviation of the posterior distribution.}
We keep $f(x)$ fixed as above: $f(x) = \arg\max_{k} \pb{k}{\sx}$.
The second confidence function we investigate is the standard deviation of the probability at $\classifier(\sx)$ under the posterior:
\begin{equation*}
  \STD^2(\sx) = \int_{\spaceparam} \pb{\classifier(\sx)}{\sx, \param}^2 \pb{\param}{\dataset} \rmd \param
  - \parenthese{\int_{\spaceparam} \pb{\classifier(\sx)}{\sx, \param} \pb{\param}{\dataset} \rmd \param }^2 \eqsp.
\end{equation*}
We estimate it by
\begin{equation}\label{eq:def-STD-hat}
  \STDhat^2(\sx) = \frac{1}{\nsamples} \sum_{i=1}^{\nsamples} \pb{\classifierhat(\sx)}{\sx, \param_i}^2
  - \parenthese{\frac{1}{\nsamples} \sum_{i=1}^{\nsamples} \pb{\classifierhat(\sx)}{\sx, \param_i}}^2 \eqsp,
\end{equation}
where $\classifierhat$ is defined in \eqref{eq:def-classifier-argmax-empirical}, and $\defEns{\param_i}_{i=1}^{\nsamples}$ are approximately drawn according to the posterior distribution.
The actual confidence function is defined as $\conf(x) = - \STD(\sx)$.

\paragraph*{Entropy of $q$.}
Finally, the last confidence measure we study is based on a probability distribution over the $\nbclass$ classes defined as
\begin{equation*}
  \pbq{k}{\sx} = \int_{\spaceparam} \1\defEns{\classifier[\param](\sx) = k} \pb{\param}{\dataset} \rmd \param \eqsp,
\end{equation*}
where $\classifier[\param](\sx) = \argmax_{k\in\defEns{1,\ldots,\nbclass}} \pb{k}{\sx, \param}$.
The idea is to measure the amount of posterior mass under which each class is selected.
The empirical estimator is given by
\begin{equation}\label{eq:def-pbqhat}
  \pbqhat{k}{\sx} = \frac{1}{\nsamples} \sum_{i=1}^{\nsamples} \1\defEns{\classifier[\param_i](\sx) = k} \eqsp.
\end{equation}
The confidence is then based on the \emph{entropy} of $\defEns{\pbq{k}{\sx}}_{k=1}^{\nbclass}$ (or $\defEns{\pbqhat{k}{\sx}}_{k=1}^{\nbclass}$, in practice): $\conf(x) = - \mathcal{H} (\pbq{\cdot}{\sx})$. 
\section{Algorithms}
\label{sec:algorithms}
In this Section, we describe a number of algorithms which allow to approximately draw samples from the posterior distribution $\param\mapsto\pb{\param}{\dataset}$. The core idea, common to all of them, consists in explicitly disentangling representation learning and uncertainty estimation.

We start by describing the high-level idea behind all the algorithms.
Let $\dataset$ be a classification training dataset.
We first train a standard deep neural network to convergence using the cross entropy loss and a classical optimizer such as Adam \cite{kingma:adam}.
We denote by $\spacez$ the vector space containing the input to the last layer of the trained neural network.
The cornerstone of our method, coming from transfer learning \cite{Yosinski:2014:TFD:2969033.2969197,Razavian:2014:CFO:2679599.2679731,pmlr-v32-donahue14}, consists first in computing the last layer features $\sz\in\spacez$ from the inputs $\sx\in\spacex$ by making a forward pass through the trained network.
We do this for all points in $\dataset$, and produce a new training dataset $\datarepr = \defEns{\sz_i, \sy_i}_{i=1}^{\ndata}$ which should provide a simpler representation of the data for the classification task.
Finally, uncertainty estimation is carried out on $\datarepr$ via any algorithm that computes confidence estimates.
In our case, the latter are applied to the last layer of the network, which is a dense layer $\param$ with a softmax activation, \ie\ for $\param = (\weights, \bias)$
\begin{equation*}
  \defEns{\pb{k}{\sz, \param}}_{k=1}^{\nbclass} = \softmax\parenthese{\weights \sz + \bias} \eqsp.
\end{equation*}
We suggest and describe below four algorithms to perform uncertainty estimation: Stochastic Gradient Descent (SGD), Stochastic Gradient Langevin Dynamics (SGLD), Monte-Carlo Dropout (MC-Dropout), and Bootstrap.
They all compute an \emph{ensemble} of models $\{ \theta_i \}_{i=1}^{\nsamples}$.
The last-layer approach is not restricted to these algorithms, and it can be implemented in combination with any algorithm computing uncertainty estimates from $\datarepr$.
Running the algorithms on the last layer considerably reduces the computational cost required to find uncertainty estimates.
Note that this two-stage procedure may make the Bayesian theory (which motivates the suggested last-layer algorithms) not hold exactly.

In \Cref{sec:sgld-sgd,sec:MC-dropout,sec:bootstrap}, we describe the last layer version of SGLD, SGD, MC-Dropout, and Bootstrap.
Recall for them the training dataset is $\datarepr$, instead of $\dataset$.
We also apply the four algorithmic ideas to the full network: adaptation is simple, by replacing $\datarepr$ by $\dataset$.
For the four algorithms, the last layer or full neural network is always initialized at $\paramstar$, the parameters of the trained network after convergence.


\subsection{Stochastic Gradient Langevin Dynamics and Stochastic Gradient Descent}
\label{sec:sgld-sgd}

Stochastic Gradient Langevin Dynamics (SGLD) is a Monte Carlo Markov Chain (MCMC) algorithm \cite{Welling:2011:BLV:3104482.3104568}, adapted from the Langevin algorithm \cite{roberts:tweedie:1996} to large-scale datasets by taking a single mini-batch of data to estimate the gradient at each update. More precisely, by the Bayes' rule, the posterior distribution $\param\mapsto\pb{\param}{\dataset}$ is proportional to $\prior{\param} \prod_{i=1}^{\ndata} \pb{\sy_i}{\sz_i, \param}$ where $\param\mapsto\prior{\param}$ is a prior distribution on $\param$.
In practice, we choose a standard Gaussian prior.
The update equation of SGLD is then given for $k\in\nset$ by
\begin{equation}\label{eq:update-sgld}
  \param_{k+1} = \param_k + \lr \parenthese{\frac{1}{\bsize}\sum_{i\in\batch} \nabla \log \pb{\sy_i}{\sz_i, \param_k} + \frac{\nabla \log \prior{\param_k}}{\ndata}}
  + \sqrt{\frac{2\lr}{\ndata}} \nZ_{k+1} \eqsp,
\end{equation}
where $\lr$ is a constant learning rate, $\batch$ a mini batch from $\datarepr$ of size $\bsize\in\nset^*$ and $(\nZ_k)_{k\in\nset^*}$ an \iid~sequence of standard Gaussian random variables of dimension $\dim \spaceparam$. Following \cite{ahn:welling:balan:2012,chen2014stochastic}, we apply SGLD with a constant learning rate.
However, a decreasing learning rate is also a valid approach.
We do not apply a burn-in period because the last layer or full neural network is always initialized at $\paramstar$, a local minima.

The update equation of SGLD \eqref{eq:update-sgld} is equal to the update equation of Stochastic Gradient Descent (SGD), apart from the addition of the Gaussian noise $\sqrt{2\lr/\ndata} \nZ$. In the same vein, \cite{Mandt:2017:SGD:3122009.3208015} shows that, under certain assumptions, SGD with a carefully chosen constant step-size can be seen as approximate sampling from a posterior distribution with an appropriate prior.
Therefore, we also consider SGD as an MCMC algorithm to approximately sample from the posterior distribution.

We apply the thinning technique to reduce the memory cost: given a thinning interval $\nthinning\in\nset^*$ and a number of samples $\nsamples\in\nset^*$, we run the Markov chain $(\param_k)_{k\in\nset}$ during $\nsamples \times \nthinning$ steps and at every $\nthinning$ iteration, we save the current parameters of the (last layer or full) neural network $\param$. The procedure is summarized in \Cref{alg:sgld-sgd}, where $\SGLD$ (resp. $\SGD$) stands for the update equation \eqref{eq:update-sgld} (resp. \eqref{eq:update-sgld} without Gaussian noise).

\begin{algorithm}[tb]
  \caption{SGLD and SGD}
  \label{alg:sgld-sgd}
\begin{algorithmic}
  \STATE {\bfseries Input:} data $\datarepr$,
  neural network $\param$,
  number of samples $\nsamples$,
  thinning interval $\nthinning$,
  batch size $\bsize$,
  learning rate $\lr$.
  \STATE {\bfseries Initialize} $\param = \param^*$.
  \FOR{$i=1$ {\bfseries to} $\nsamples$}
    \FOR{$j=1$ {\bfseries to} $\nthinning$}
      \STATE $\param \leftarrow \SGLD(\param, \lr, \bsize) \text{ or } \SGD(\param, \lr, \bsize)$
    \ENDFOR
    \STATE {\bfseries Save} $\param$.
  \ENDFOR
\end{algorithmic}
\end{algorithm}

\subsection{Monte-Carlo Dropout}
\label{sec:MC-dropout}

Dropout provides a popular method for computing empirical uncertainty estimates, and it was initially developed to avoid over-fitting in deep learning models \cite{JMLR:v15:srivastava14a}.
It approximately samples from the posterior distribution $\param\mapsto\pb{\param}{\dataset}$ when applied at test time \cite{Gal:2016:DBA:3045390.3045502}.
This technique, often named Monte-Carlo Dropout (MC-Dropout), is widely used in practical applications \cite{Zhu2017DeepAC,leibig2017leveraging,Nair2018ExploringUM} due to its simplicity and good performance.

Dropout randomly sets a fraction $\pdrop\in\ooint{0,1}$ of input units to 0 at
each update during training time, or at each forward pass during test time.
For the full network version, we interleave a dropout layer after each max
pooling layer in the VGG-type neural network and before each dense layer. The
method is described in \Cref{alg:mc-dropout}. 

\begin{algorithm}[tb]
   \caption{MC-Dropout}
   \label{alg:mc-dropout}
\begin{algorithmic}
  \STATE {\bfseries Input:} data $\datarepr$,
  neural network $\param$,
  number of samples $\nsamples$,
  dropout rate $\pdrop$,
  batch size $\bsize$,
  learning rate $\lr$,
  number of training epochs $\nepochs$.
  \STATE {\bfseries Initialize} $\param = \param^*$.
  \STATE {\bfseries Train} $\param$, using SGD with a learning rate $\lr$, batch size $\bsize$, dropout rate $\pdrop$ and a number of epochs $\nepochs$.
  \STATE {\bfseries Save} $\param$.
  \STATE For a given input $x$, we run $ \nsamples$ forward passes from $\param$ using dropout again.
\end{algorithmic}
\end{algorithm}

\subsection{Bootstrap}
\label{sec:bootstrap}

At the crossroad between the Bayesian and the frequentist approaches, the Bootstrap algorithm may provide a simple way to approximate the sampling distribution of an estimator, see \eg~\cite{Efron2012,efron:aas:2012}.
We first sample with replacement $\ndata$ data points from the training dataset $\datarepr$, thus generating a new bootstrapped dataset $\databoot$.
After this, either only the last layer (multinomial logistic regression) or a full neural network is trained on $\databoot$ until convergence, and the parameters of the network $\param$ are saved.
We repeat this as many times as models we want, and then compute their ensemble.
The procedure is detailed in \Cref{alg:bootstrap}.

\begin{algorithm}[tb]
   \caption{Bootstrap}
   \label{alg:bootstrap}
\begin{algorithmic}
  \STATE {\bfseries Input:} data $\datarepr$,
  neural network $\param$,
  number of samples $\nsamples$,
  batch size $\bsize$,
  learning rate $\lr$,
  number of training epochs $\nepochs$.
  \FOR{$i=1$ {\bfseries to} $\nsamples$}
    \STATE {\bfseries Initialize} $\param = \param^*$.
    \STATE {\bfseries Sample} a bootstrapped dataset $\databoot$ from $\datarepr$.
    \STATE {\bfseries Train} $\param$ on $\databoot$, using SGD with a learning rate $\lr$, batch size $\bsize$ and a number of epochs $\nepochs$.
    \STATE {\bfseries Save} $\param$.
  \ENDFOR
\end{algorithmic}
\end{algorithm}

\section{Experimental Results and Discussion}
\label{sec:experiments}

We evaluate the quality of the uncertainty estimates produced by the last-layer algorithms on four image classification tasks of increasing complexity.
The MNIST dataset \cite{726791} consists of 28x28 handwritten digits, which are divided in a training set with 60000 examples and a test set with 10000 images.
The CIFAR-10 (resp. CIFAR-100) datasets \cite{Krizhevsky09learningmultiple} consists of 32x32x3 colour images, each one corresponding to one of 10 (resp. 100) classes.
The dataset is split in 50000 training images, and 10000 test ones.
Therefore, there are 6000 (resp. 600) images per class.
Finally, the ImageNet dataset \cite{imagenet_cvpr09} has 1281167 training images and 50000 test ones, and they are divided in 1000 classes. We randomly crop the colour images to a 331x331x3 size.

For MNIST, we consider a fully-connected feedforward neural network with 2 hidden layers of 512 and 20 neurons respectively. For CIFAR-10 and 100, we use a pretrained VGG-16 neural network\footnote{\url{https://github.com/geifmany/cifar-vgg}} with 512 neurons in the last hidden layer. For ImageNet, the 4032-dimensional last-layer features are computed through a pretrained NASNet neural network\footnote{\url{https://keras.io/applications/\#nasnet}}.
The trained networks achieve a standard accuracy of $98\%$ for MNIST, $94\%$ for CIFAR-10, $70\%$ for CIFAR-100, and $69\%$ for ImageNet (top-1 accuracy).
See \Cref{table:accuracies} in the appendix for test accuracies for all algorithms and datasets.

In addition to the four algorithms MC-Dropout, Bootstrap, SGD and SGLD, we evaluate the SGD-Point Estimate (SGD-PE) baseline which simply computes the softmax outputs provided by the pretrained neural network. The posterior approximation is then formally a Dirac at $\paramstar$, the parameters of the pretrained network. Thus, the only confidence function we can compute for SGD-PE is the softmax response $\SR$ or its empirical estimation $\SRhat$ defined in \eqref{eq:def-SR-empirical}.

We conduct two sets of experiments: we first evaluate the five methods against the AURC metric and then their ability to detect out-of-distribution samples (AUROC and AUPR-in/out). The results for the latter are reported in \Cref{sec:suppl-out-of-distribution}.
In order to better understand the value of adding multiple uncertainty layers, we run the algorithms both on the last layer and on the full neural networks for MNIST and CIFAR-10/100. We append the word \textsc{full} to denote the full network versions of the algorithms in the tables below.

Given the size of both ImageNet and the NASNet network, we assess the potential benefit of multiple uncertainty layers on ImageNet by adding up to 3 dense hidden layers with 4032 neurons on top of NASNet.
We apply the uncertainty algorithms to one, two, or the three layers.
For example, in the case of Dropout, we compare the performance of adding from one to three dropout layers (note we do \emph{not} add any layers without dropout in this case).
For control, we also run SGD-PE in the three fully-connected architectures.
The dense layers added at the top of NASnet are fine-tuned first, and these weights are then used both as a reference (SGD-PE) and as the starting point $\paramstar$ for the four algorithms.

We perform a hyper-parameter search for all algorithms and datasets.
Details are in \Cref{sec:additional-experiments}.
We only report below results for the best hyper-parameter values.


\begin{figure}[] 
\begin{center}
\includegraphics[scale=0.4]{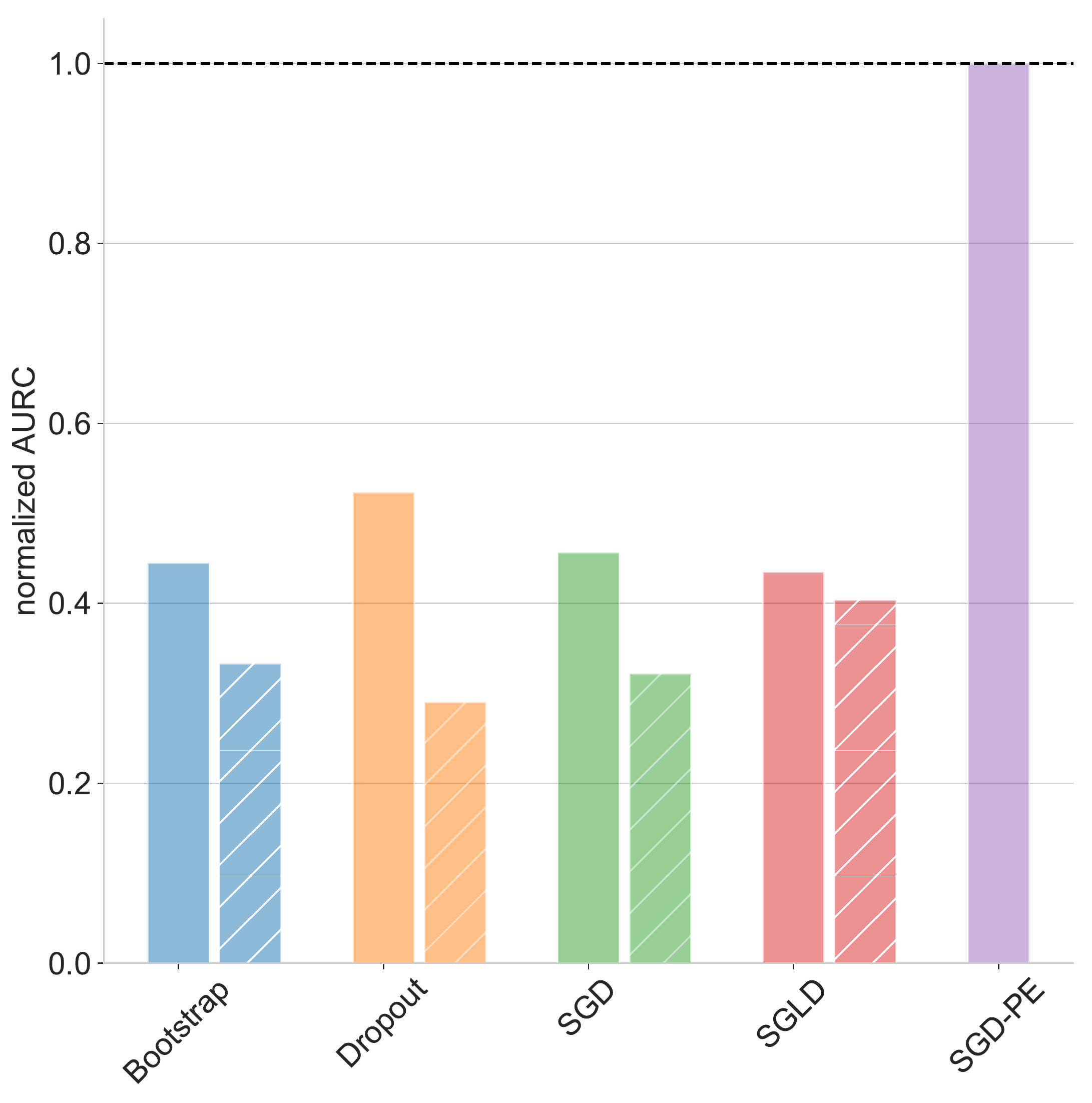}
\caption{
Normalized AURC for last-layer (solid) and full network (striped) versions of the four algorithms: Bootstrap, MC-Dropout, SGD, SGLD, and SGD-PE baseline, on \textbf{MNIST}.
}
\label{figure:aurc-mnist}
\end{center}
\end{figure}


\begin{figure}[] 
\begin{center}
\includegraphics[scale=0.3]{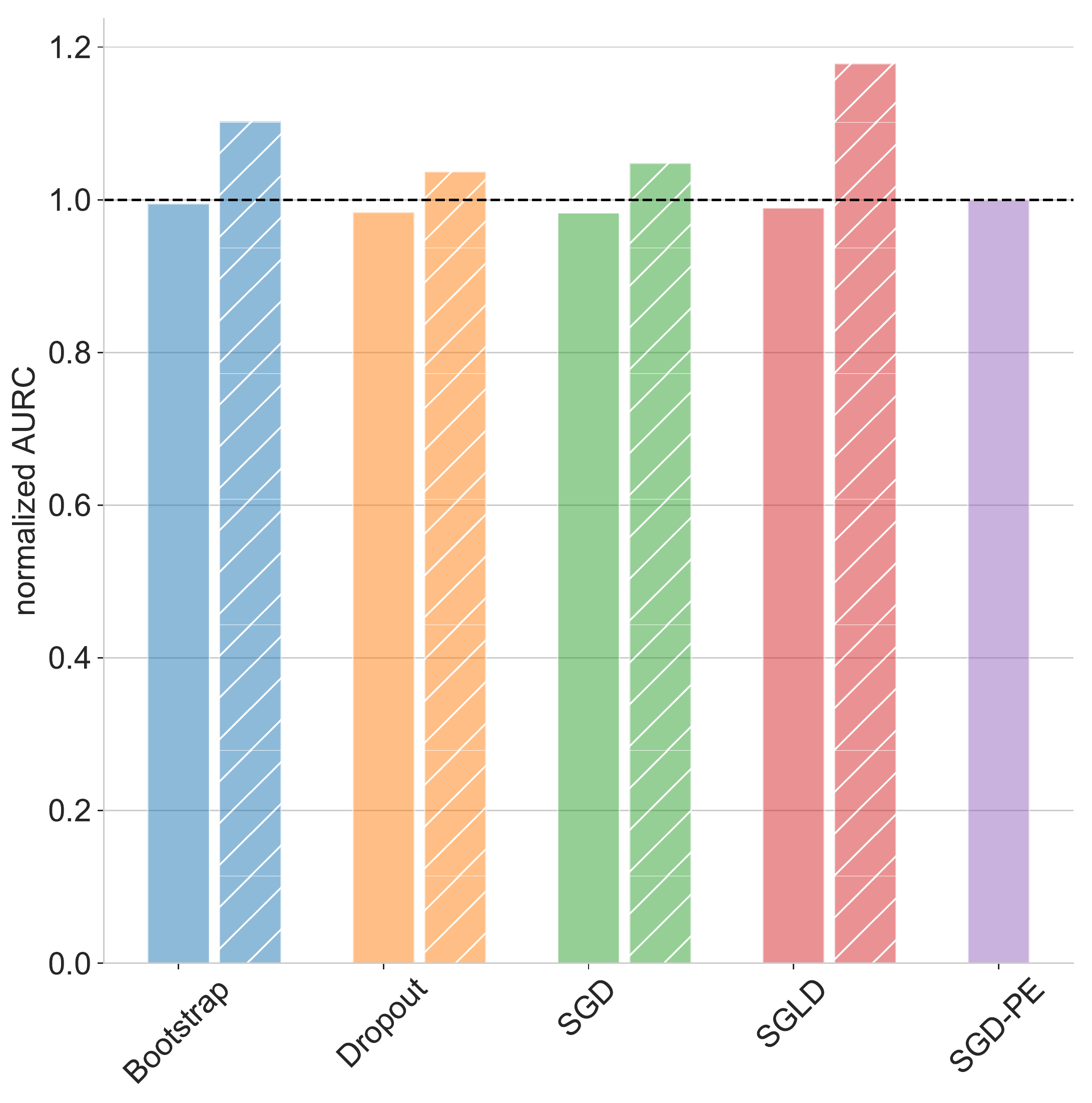}
\caption{
Normalized AURC for last-layer (solid) and full network (striped) versions of the four algorithms: Bootstrap, MC-Dropout, SGD, SGLD, and SGD-PE baseline, on \textbf{CIFAR-100}.
}
\label{figure:aurc-cifar100}
\end{center}
\end{figure}

\begin{figure}[] 
\begin{center}
\includegraphics[scale=0.3]{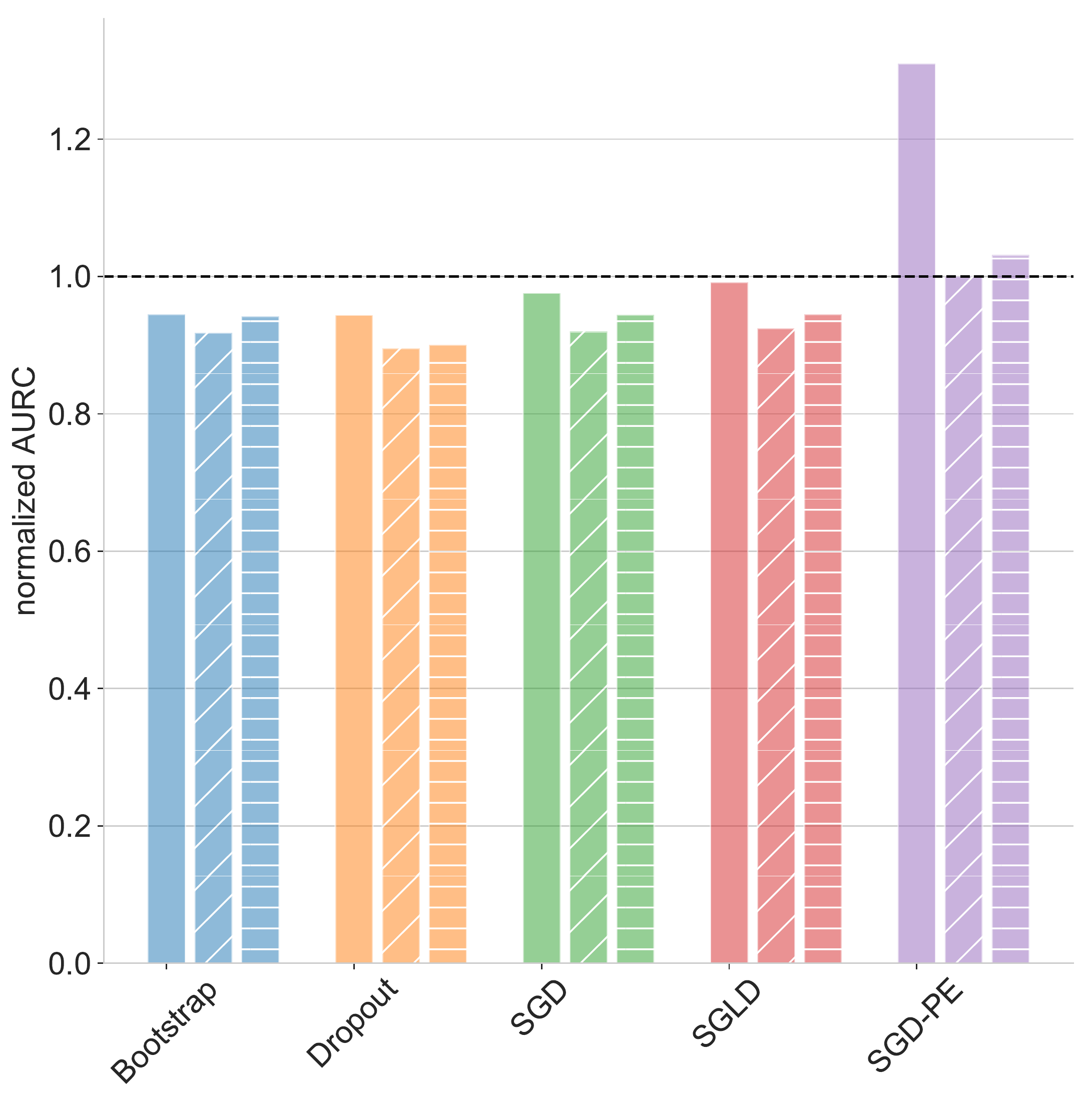}
\caption{
Normalized AURC for the 1 (solid), 2 (45-degree stripes) to 3 (horizontally striped) dense layer(s) versions of the algorithms: Bootstrap, MC-Dropout, SGD, SGLD and SGD-PE baseline, on \textbf{ImageNet}.
The normalized AURC is based on the AURC obtained using SGD-PE on 2 dense layers on top of NASNet.
}
\label{figure:aurc-imagenet}
\end{center}
\end{figure}

The results for the AURC metric are shown in \Cref{figure:aurc-mnist} for MNIST, while \Cref{figure:aurc-cifar100} contains the outcomes for CIFAR-100, and \Cref{figure:aurc-imagenet} those for ImageNet.
We recall that the lower is the AURC, the better is the result.
Let us define by min AURC the minimum value achieved using either SR, STD or the entropy of $\hat{q}$ as a confidence function.
For better readability, we define the normalized AURC as the ratio of min AURC over the AURC of SGD-PE (unique, using SR as confidence function).
Tables are provided in \Cref{sec:suppl-selective-classification}.
Note that our results are reported for one run of the algorithms because Bayesian approaches include uncertainty estimates.

We would like to pursue several avenues of research in the future: first, to compare our methodology with ensemble methods over the full network, \ie~snapshot ensembles \cite{2017arXiv170400109H}, deep ensembles \cite{NIPS2017_7219} as well as other methods that can be described as being Bayesian about the last layer such as deep kernel learning \cite{pmlr-v51-wilson16}. Second, to make additional comparisons to temperature scaling \cite{guo:pleiss:sun:weinberger:2017} and to apply bootstrapping to the full dataset and training procedure.
Third, to consider methods for variational inference and Laplace approximations, \eg~\cite{ritter2018a}, and alternative methods for Bayesian logistic regression on large datasets since the focus is on last layer Bayesian approaches using the features learned from deep neural networks, \eg~\cite{NIPS2016_6486,pmlr-v38-hoffman15}.

We summarize the results we obtain as follows:

\begin{figure}[] 
\begin{center}
\includegraphics[scale=0.5]{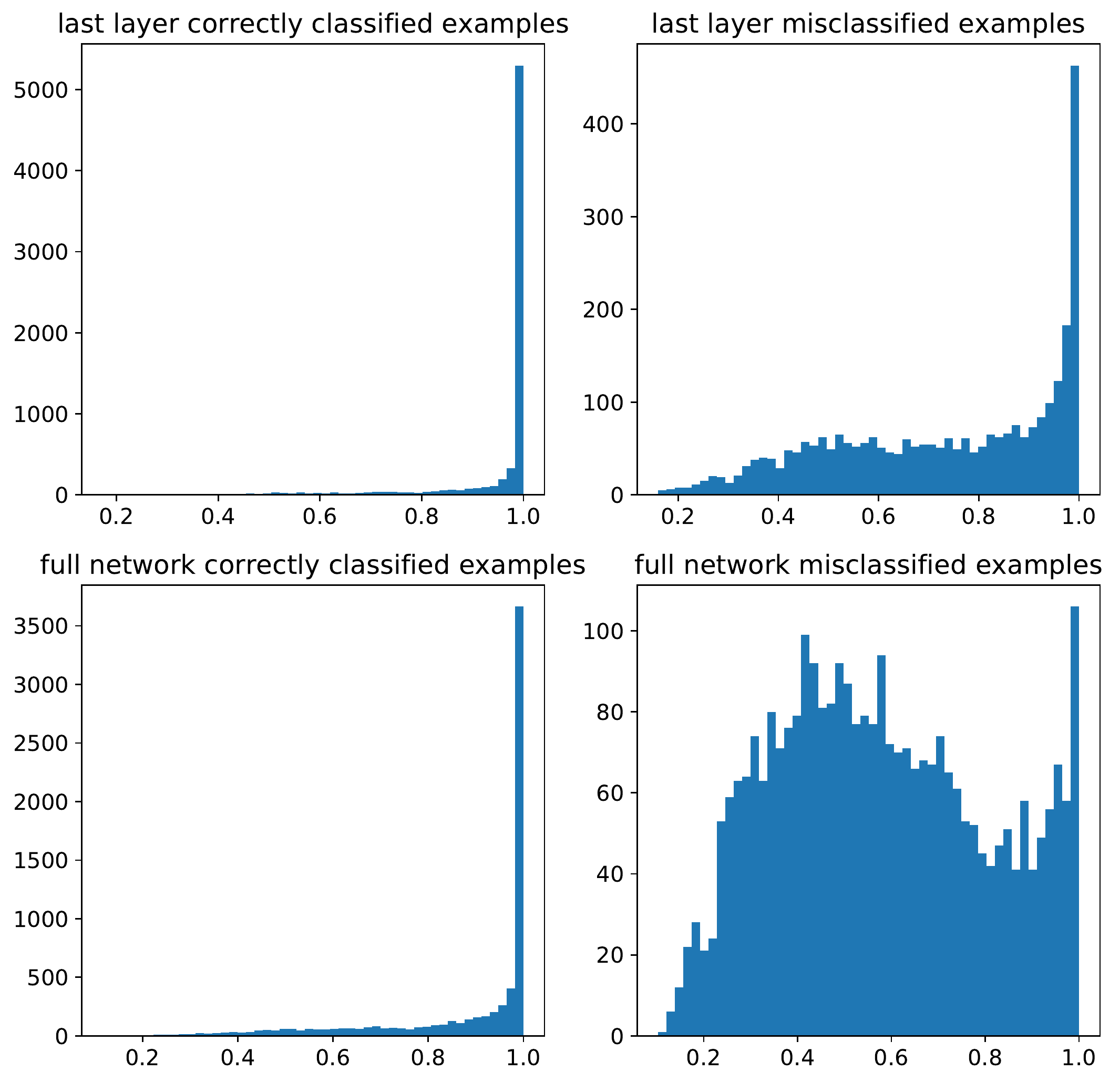}
\caption{
Histograms of the $\SRhat$ confidence function values defined in \eqref{eq:def-SR-empirical} for all correctly classified and misclassified test data points.
$x$-axis: $\SRhat$ values. $y$-axis: frequency per bin.
\textbf{Top row:} Last layer version of SGD on CIFAR-100.
\textbf{Bottom row:} Full network version of SGD on CIFAR-100.
\textbf{Left column:} Correctly classified test data points.
\textbf{Right column:} Misclassified test data points.
}
\label{figure:cifar100_pmax_hist_sgd}
\end{center}
\end{figure}

\begin{figure}[] 
\begin{center}
\includegraphics[scale=0.5]{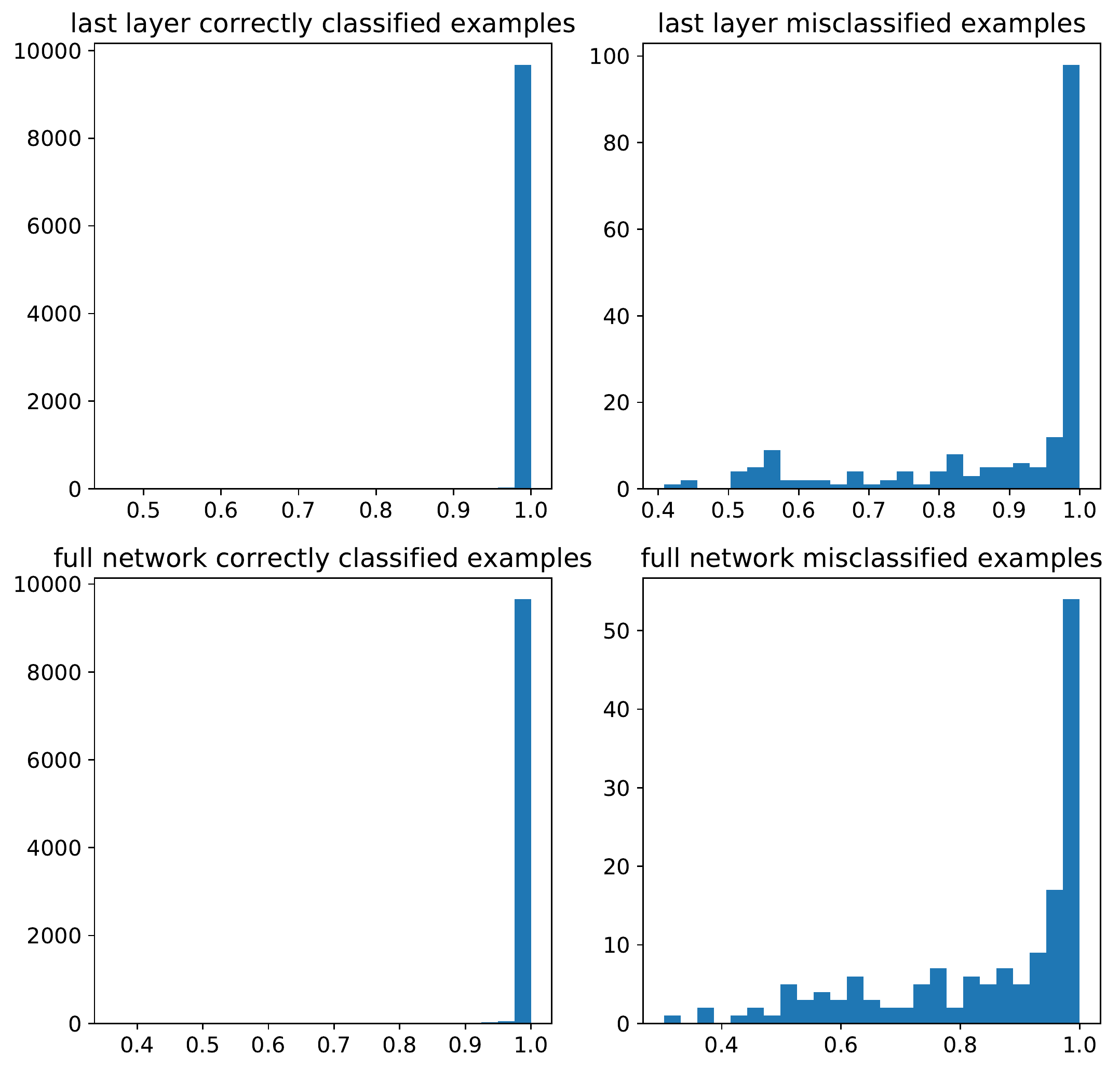}
\caption{Similar plot as \Cref{figure:cifar100_pmax_hist_sgd}, but for the MNIST dataset.}
\label{figure:mnist_pmax_hist_sgd}
\end{center}
\end{figure}

\paragraph*{1) Adding Multiple Uncertainty Layers Does Not Help.}
Except on the MNIST dataset, where adding an extra hidden uncertainty layer improves the AURC, the last layer and its full network counterpart seem to perform similarly well for the four algorithms.
On CIFAR-10, AURC is actually better for the last layer algorithms: $9\%$ better for Dropout, $8\%$ for Bootstrap, $2\%$ for SGD and $14\%$ for SGLD. A similar observation can be made about CIFAR-100: $6\%$ better for Dropout, $11\%$ for Bootstrap, $7\%$ for SGD and $19\%$ for SGLD.
AURC is mostly constant with respect to the number of uncertainty layers (from 1 to 3) for ImageNet; a maximum variation of $7\%$ can be observed. More precisely, the best performance for all algorithms is obtained with $2$ uncertainty layers.

We show histograms of the SR confidence function values to shed some light on the difference between MNIST and CIFAR-10/100. See \Cref{figure:cifar100_pmax_hist_sgd} for CIFAR-100 and \Cref{figure:mnist_pmax_hist_sgd} for MNIST.
These plots compare the SR distributions of the correctly classified and misclassified test points for last-layer SGD versus its full network counterpart. Actually, AURC is a direct measure of how well the two distributions are separated from each other: in the extreme case where all the correctly classified points had higher SR values than the misclassified ones, AURC would reach its minimum value (known as E-AURC in \cite{geifman:uziel:yaniv:2018}).

In the case of MNIST, the histograms for correctly classified points are similar for both last-layer and full-network SGD versions.
However, the full-network exhibits a greater dispersion for incorrectly classified points (see scale of y-axis). Both facts combined lead to a stronger AURC for the full-network algorithm, as it can better tell the difference between both sets of points.
Indeed, MNIST is known to be an easy classification task, and a flat landscape is to be expected for local loss minima. In other words, there are many possible distinct representations which are enough to solve the problem. In this case, full-network methods are able to explore many of them, thus accounting for the uncertainty in the representation, while still correctly classifying with high confidence the vast majority of points.
Accordingly, we suspect full-network approaches provide a more diverse set of predictions in this context, when compared to the last layer implementation which is committed to a single representation.
We believe that, when the loss landscape is more-or-less flat, full-network algorithms can take advantage of representation uncertainty to deliver stronger results.


A different behavior can be observed on CIFAR-100, where the classification task
is more difficult. The histograms of the full-network SGD are more dispersed for
\emph{both} correctly classified and misclassified points. In particular, as
opposed to the MNIST scenario, a number of correctly classified points are no
longer mapped to a high SR. Therefore, the confidence function SR is worse at
ranking these examples, while the effect is lighter for the last layer version
of the network, leading to a better AURC in the latter case. One possible
explanation could be related to the \emph{sensitivity} of the representation
found by the pretrained network. For hard classification problems (like
CIFAR-100 or ImageNet), the network is expected to end up in a strong or deep
local minima after training. Intuitively, this means that the quality of nearby
representations in $\theta$-space quickly degrades. Unfortunately, this is
precisely what most of the full-network uncertainty methods try to do: they
apply some dithering to the original strong local minima. A number of previously
correctly \emph{ranked} points might suffer due to poor representations. In this
case, committing to the local optima may pay off; last-layer models still
exploit the fixed representation to compute some useful uncertainty estimates on
top.
In difficult problems, when we bootstrap the data, each individual model gets exposed to fewer different data points.
We suspect this leads to reaching worse local minima, than the aforementioned deep one.
One could try to train an ensemble of networks on the very same data.
This may help in practice for this type of hard problems.
It could also be the case that, if many of those models end up in the same deep local minima, their predictions will not be diverse enough to generate useful uncertainty estimates.
In these cases, we expect last-layer models to help at a reasonable computational cost.
We see that the AURC is slightly worse in CIFAR-100 for the full network methods (between $6\%$ and $19\%$).
Therefore, for harder classification tasks, our results seem to support the idea that by explicitly decoupling representation learning (based on all but one layers) from uncertainty estimation (which is fully performed at the last layer) we capture most of the value provided by these algorithmic approaches in terms of selective classification.

\paragraph*{2) Softmax Response (SR) is a Strong Confidence Function.}
We have compared several confidence functions: SR, STD and the entropy of $\hat{q}$ defined in \Cref{sec:confidence}.
We observe a common theme in all cases: the softmax response SR does consistently outperform all the other confidence functions.
As an example, the risk coverage curve for ImageNet is plotted in \Cref{figure:imagenet_risk_coverage} of the appendix.

\paragraph*{3) SGD Point-Estimate is actually a Strong Baseline.}
SGD-PE is particularly competitive on CIFAR-10/100; it provides almost optimal performance.
Its main advantage is simplicity: it can be applied off-the-shelf and no two-stage procedure is needed.
However, the method suffers in both MNIST and ImageNet, compared to the other algorithms.
For MNIST, the explanation is similar to the previous discussion: MNIST being an easy classification task, the full network versions of our algorithm are superior to the last layer versions, which are themselves better than a single point estimate.
For ImageNet, the results suggest that raw end-to-end softmax outputs may not be enough for more complex datasets. \emph{Ensemble} techniques may bring additional stability and robustness in this context.

\paragraph*{4) SGLD is Unstable on the Full Network.}
When running SGLD on the full network for MNIST and CIFAR-10/100, we observed the instability of this algorithm: if the learning rate is not very small, SGLD tends to diverge, \ie~the accuracy (resp. the loss) decreases (resp. increases) over the iterations. This phenomenon is not visible when SGLD is only applied on the \emph{last layer} of the neural network.
In the case of one dense layer endowed with a multinomial logistic regression model and a Gaussian prior over the weights and bias, the logarithm of the posterior distribution $\param\mapsto\pb{\param}{\dataset}$ is a \emph{strongly log concave} function.
In this setting, convergence properties of SGLD have been studied \cite{dalalyan:2017,2017arXiv171000095D} and its behaviour has been shown to be close to SGD \cite{2017arXiv170602692N,NIPS2018_8048}.
We conclude the full network version of SGLD should be used carefully while its last-layer counterpart should be easier to train.
In the future, we intend to perform comparisons with a decreasing learning rate schedule and make tests of convergence of the SGLD chain, \eg~number of effective samples.

The results for out-of-distribution detection are in \Cref{sec:suppl-auroc-aupr-results}. They support similar take-away messages.

\section{Conclusion}
\label{sec:conclusion} 
In this work, we showed that decoupling representation learning and uncertainty
quantification in deep neural nets is a tractable approach to tackle selective
classification, which is an important problem for real-world applications where
mistakes can be fatal. Vanilla methods that do not compute uncertainty estimates
struggle to solve some of the most complex tasks we studied. In addition, our
experiments indicate that the improvements obtained by adding several
uncertainty layers (either at the top, or along the whole architecture) are at
most modest, thus making it hard to justify their complexity overhead.

\appendix

\clearpage

\section*{Appendix}

In \Cref{sec:additional-experiments}, additional material concerning the experiments is provided: the hyper-parameters tuning is detailed and supplementary tables and plots for the AURC metric are presented.
In \Cref{sec:suppl-out-of-distribution}, metrics for out-of-distribution detection are first defined and then computed for the last layer algorithms on MNIST and CIFAR-10/100: related tables and plots are displayed and presented.

\section{Additional Material for the Experiments}
\label{sec:additional-experiments}

The accuracies for all algorithms and all datasets are presented in \Cref{table:accuracies}.

\begin{table}[tb]
	\caption{Accuracies for all algorithms and datasets.}
	\label{table:accuracies}
	\vskip 0.15in
	\begin{center}
		\begin{small}
			\begin{sc}
				\begin{tabular}{l|ccccc}
					\toprule
                & mnist & cifar10 & cifar100 \\
                     \midrule
sgd             & 0,981 & 0,936   & 0,706    \\
sgld            & 0,981 & 0,935   & 0,707    \\
bootstrap       & 0,981 & 0,935   & 0,704    \\
dropout         & 0,980 & 0,935   & 0,705    \\
sgd-pe          & 0,978 & 0,936   & 0,705    \\
sgd full       & 0,985 & 0,933   & 0,696    \\
sgld full      & 0,982 & 0,929   & 0,677    \\
bootstrap full & 0,984 & 0,931   & 0,687    \\
dropout full   & 0,984 & 0,931   & 0,700    \\
					\bottomrule
				\end{tabular}
			\end{sc}
		\end{small}
	\end{center}
	\vskip -0.1in
\end{table}

\subsection{Hyper-Parameter Tuning}
\label{sec:hyper-params}

We perform hyper-parameter tuning for the last layer algorithms as follows. \\
\textbf{Algorithms}. The number of samples $\nsamples$ varies between 10, 100, and 1000 for SGD, SGLD and MC-Dropout and between 10 and 100 for Bootstrap.
The thinning interval $\nthinning$ is chosen such that the parameters $\param$ are saved at each epoch (one full pass over the data) for SGLD and SGD. 
The number of SGD epochs $\nepochs$ is 10 for the Bootstrap, and 100 for MC-Dropout.
The dropout rate $\pdrop$ for the latter (probability of zeroing out a neuron) is chosen between 0.1, 0.3, and 0.5. \\
\textbf{MNIST}. The learning rate $\lr$ is among 5 equally spaced values between $10^{-1}$ and $10^{-3}$. We use batch-size $\bsize$ equal to 32. \\
\textbf{CIFAR-10/100}. The learning rate is among 7 equally spaced values between $10^{-2}$ and $10^{-5}$. The batch-size is 128. \\
\textbf{ImageNet}. The learning rate is among 4 equally spaced values between $10^{-1}$ and $10^{-4}$, with a batch-size of 512. In this case, the number of samples $\nsamples$ is 10, and only 10 epochs are completed for MC-Dropout.

For the full network versions of the algorithms on MNIST and CIFAR-10/100, the number of samples $\nsamples$ is equal to $100$, the number of epochs $\nepochs$ is 10 for Bootstrap and $100$ for MC-Dropout. \\
\textbf{MNIST}. The learning rate is among 4 equally spaced values between $10^{-1}$ and $10^{-4}$. We use batch-size equal to 32. \\
\textbf{CIFAR-10/100}. The learning rate is among 4 equally spaced values between $10^{-2}$ and $10^{-5}$. The batch-size is 128.
The metrics optimized by the hyper-parameter search are defined in \Cref{sec:metrics}.

\subsection{Additional Results for Selective Classification}
\label{sec:suppl-selective-classification}

Tables for the AURC metric are shown in \Cref{table:mnist-aurc} for MNIST, while \Cref{table:cifar-10-aurc} contains the outcomes for CIFAR-10, and \Cref{table:cifar-100-aurc} those for CIFAR-100.
Finally, ImageNet results are displayed in \Cref{table:imagenet-aurc}.
\textsc{AURC sr} (resp. \textsc{AURC std}) is the AURC obtained when the confidence function is the softmax response SR (resp. STD). \textsc{min AURC} is the minimum of these two values, and \textsc{increase} is the ratio of the min AURC over the AURC obtained by SGD-PE.
For ImageNet, the ratio is over the AURC using SGD-PE on a 2 dense layers network at the top of NASNet.
The AURC using the entropy of $\hat{q}$ defined in \eqref{eq:def-pbqhat} as a confidence function is not reported because the results are clearly below its competitors. In \Cref{table:imagenet-aurc}, \textsc{nb-ll} indicates the number of dense layers added at the top of NASNet (from 1 to 3).

\begin{table*}[tb]
	\caption{AURC for MC-Dropout, Bootstrap, SGD, SGLD and SGD-PE on the MNIST dataset.}
	\label{table:mnist-aurc}
	\vskip 0.15in
	\begin{center}
		\begin{small}
			\begin{sc}
				\begin{tabular}{|l|ccc|c|r}
					\toprule
					algorithm & AURC sr & AURC std & min AURC & increase \\
					\midrule
					dropout         & 8.74E-04      & 1.10E-03  & 8.74E-04  & 0.52         \\
					\textbf{dropout full}   & \textbf{4.84E-04} & \textbf{5.57E-04} & \textbf{4.84E-04} & \textbf{0.29}         \\
					bootstrap       & 7.43E-04      & 7.55E-04  & 7.43E-04  & 0.45         \\
					bootstrap full & 7.68E-04      & 5.56E-04  & 5.56E-04  & 0.33         \\
					sgd             & 1.18E-03      & 7.62E-04  & 7.62E-04  & 0.46         \\
					sgd full       & 5.78E-04      & 5.37E-04  & 5.37E-04  & 0.32         \\
					sgld            & 7.26E-04      & 7.28E-04  & 7.26E-04  & 0.44         \\
					sgld full      & 9.03E-04      & 6.74E-04  & 6.74E-04  & 0.40         \\
					\midrule
					sgd-pe          & 1.67E-03      &           & 1.67E-03  & 1.00         \\
					\bottomrule
				\end{tabular}
			\end{sc}
		\end{small}
	\end{center}
	\vskip -0.1in
\end{table*}

\begin{table*}[tb]
	\caption{AURC for MC-Dropout, Bootstrap, SGD, SGLD and SGD-PE on the CIFAR-10 dataset.}
	\label{table:cifar-10-aurc}
	\vskip 0.15in
	\begin{center}
		\begin{small}
			\begin{sc}
				\begin{tabular}{l|ccc|c|r}
					\toprule
					algorithm & AURC sr & AURC std & min AURC & increase \\
					\midrule
					dropout         & 6.56E-03      & 6.66E-03  & 6.56E-03  & 0.98         \\
					dropout full   & 7.12E-03      & 7.32E-03  & 7.12E-03  & 1.07         \\
					bootstrap       & 6.60E-03      & 6.90E-03  & 6.60E-03  & 0.99        \\
					bootstrap full & 7.14E-03      & 7.26E-03  & 7.14E-03  & 1.07         \\
					sgd             & 6.56E-03      & 7.30E-03  & 6.56E-03  & 0.98         \\
					sgd full       & 6.69E-03      & 7.02E-03  & 6.69E-03  & 1.00         \\
					\textbf{sgld}   & \textbf{6.51E-03} & \textbf{6.80E-03} & \textbf{6.51E-03}  & \textbf{0.98}         \\
					sgld full      & 7.44E-03      & 7.56E-03  & 7.44E-03  & 1.12         \\
					\midrule
					sgd-pe          & 6.66E-03      &           & 6.66E-03  & 1.00         \\
					\bottomrule
				\end{tabular}
			\end{sc}
		\end{small}
	\end{center}
	\vskip -0.1in
\end{table*}

\begin{table*}[tb]
	\caption{AURC for MC-Dropout, Bootstrap, SGD, SGLD and SGD-PE on the CIFAR-100 dataset.}
	\label{table:cifar-100-aurc}
	\vskip 0.15in
	\begin{center}
		\begin{small}
			\begin{sc}
				\begin{tabular}{|l|ccc|c|r}
					\toprule
					algorithm & AURC sr & AURC std & min AURC & increase \\
					\midrule
					\textbf{dropout}         & \textbf{9.10E-02}      & \textbf{9.31E-02}  & \textbf{9.10E-02}  & \textbf{0.98}         \\
					dropout full   & 9.59E-02      & 1.12E-01  & 9.59E-02  & 1.04         \\
					bootstrap       & 9.20E-02      & 9.90E-02  & 9.20E-02  & 0.99         \\
					bootstrap full & 1.02E-01      & 1.14E-01  & 1.02E-01  & 1.10         \\
					\textbf{sgd} & \textbf{9.09E-02} & \textbf{9.79E-02}  & \textbf{9.09E-02}  & \textbf{0.98}         \\
					sgd full       & 9.69E-02      & 1.09E-01  & 9.69E-02  & 1.05         \\
					sgld            & 9.15E-02      & 9.45E-02  & 9.15E-02  & 0.99         \\
					sgld full      & 1.09E-01      & 1.13E-01  & 1.09E-01  & 1.18         \\
					\midrule
					sgd-pe          & 9.25E-02      &           & 9.25E-02  & 1.00         \\
					\bottomrule
				\end{tabular}
			\end{sc}
		\end{small}
	\end{center}
	\vskip -0.1in
\end{table*}

\begin{table*}[tb]
\caption{AURC for MC-Dropout, Bootstrap, SGD, SGLD and SGD-PE on the ImageNet dataset.}
\label{table:imagenet-aurc}
\vskip 0.15in
\begin{center}
\begin{small}
\begin{sc}
\begin{tabular}{|l|c|ccc|c|r}
\toprule
algorithm & nb-ll & AURC sr & AURC std & min AURC & increase \\
\midrule
\textbf{dropout}   & \textbf{1}     & \textbf{0.0974}        & \textbf{0.1318}    & \textbf{0.0974}    & \textbf{0.94}           \\
bootstrap & 1     & 0.0975        & 0.1188    & 0.0975    & 0.94           \\
sgd       & 1     & 0.1007        & 0.2569    & 0.1007    & 0.98           \\
sgld      & 1     & 0.1023        & 0.1740    & 0.1023    & 0.99           \\
sgd-pe  & 1     & 0.1352        &           & 0.1352    & 1.31           \\
\midrule
\textbf{dropout}   & \textbf{2}     & \textbf{0.0924}        & \textbf{0.1117}    & \textbf{0.0924}    & \textbf{0.90}           \\
bootstrap & 2     & 0.0947        & 0.1073    & 0.0947    & 0.92           \\
sgd       & 2     & 0.0949        & 0.1080    & 0.0949    & 0.92           \\
sgld      & 2     & 0.0954        & 0.1101    & 0.0954    & 0.92           \\
sgd-pe  & 2     & 0.1032        &           & 0.1032    & 1.00           \\
\midrule
\textbf{dropout}   & \textbf{3}     & \textbf{0.0929}        & \textbf{0.1186}    & \textbf{0.0929}    & \textbf{0.90}           \\
bootstrap & 3     & 0.0972        & 0.1102    & 0.0972    & 0.94           \\
sgd       & 3     & 0.0974        & 0.1159    & 0.0974    & 0.94           \\
sgld      & 3     & 0.0975        & 0.1119    & 0.0975    & 0.94           \\
sgd-pe  & 3     & 0.1064        &           & 0.1064    & 1.03 \\
\bottomrule
\end{tabular}
\end{sc}
\end{small}
\end{center}
\vskip -0.1in
\end{table*}

\begin{figure}[tb]
\begin{center}
\includegraphics[scale=0.5]{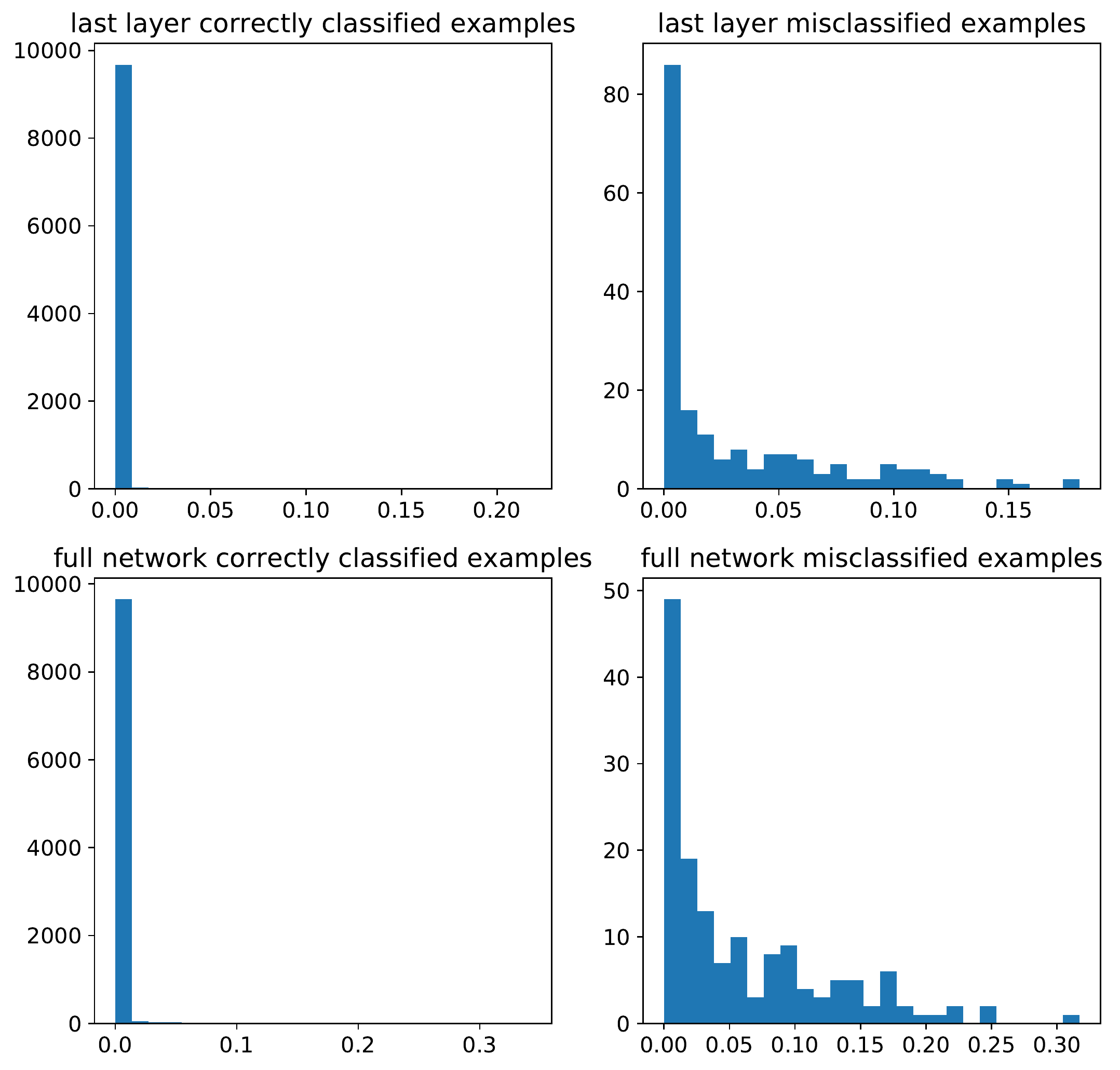}
\caption{
Histograms of the values of the $\STDhat$ confidence function defined in \eqref{eq:def-STD-hat} for the correctly classified and misclassified test examples.
horizontal axis: $\STDhat$ values. vertical axis: number of items in each bin.
\textbf{Top row:} Last layer version of SGD on MNIST. Left: histograms for the correctly classified examples. Right: for the misclassified examples.
\textbf{Bottom row:} Full network version of SGD on MNIST.
}
\label{figure:mnist_std_hist_sgd}
\end{center}
\end{figure}

\Cref{figure:mnist_std_hist_sgd} follows the same approach as \Cref{figure:cifar100_pmax_hist_sgd,figure:mnist_pmax_hist_sgd} of the article, at the exception that the confidence function is $\STDhat$ and not $\SRhat$. We observe the same phenomenon as in \Cref{figure:mnist_pmax_hist_sgd}: the histogram of the full network version of SGD is more dispersed on the misclassified examples but not on the correctly classified examples.

Already discussed in \Cref{sec:metrics} of the article, \emph{calibration} is a desired property of probabilistic models. Before computing some metrics associated to calibration, we first introduce some notations and concepts.
Our probabilistic model is said to be calibrated if for all $\alpha\in\ccint{0,1}$,
\begin{equation}\label{eq:calibration-1}
  \probacond{\sY = \argmax_{k\in\defEns{1,\ldots,\nbclass}} \pb{k}{\sX}}{\max_{k\in\defEns{1,\ldots,\nbclass}} \pb{k}{\sX} = \alpha} = \alpha \eqsp.
\end{equation}
where $\defEns{\pb{k}{\sx}}_{k=1}^{\nbclass}$ is the predictive posterior distribution defined in \eqref{eq:def-predictive-posterior}.
The empirical equivalent of \eqref{eq:calibration-1} over the test dataset $\datatest$ is
\begin{equation}\label{eq:calibration-2}
  \sum_{i=1}^{\ntest} \1\bigg\{\sy_i = \argmax_{k\in\defEns{1,\ldots,\nbclass}} \pbhat{k}{\sx_i}, \max_{k\in\defEns{1,\ldots,\nbclass}} \pbhat{k}{\sx_i} = \alpha \bigg\}  = \alpha \times \sum_{i=1}^{\ntest} \1\defEns{\max_{k\in\defEns{1,\ldots,\nbclass}} \pbhat{k}{\sx_i} = \alpha} \eqsp.
\end{equation}
In practice, it is necessary to discretize $\alpha\in\ccint{0,1}$ in $m\in\nset^*$ bins, $\coint{\alpha_0,\alpha_1}, \ldots, \coint{\alpha_{m-1}, \alpha_m}$ where $\alpha_0=0<\alpha_1<\ldots<\alpha_{m-1}<\alpha_m=1$. For $j\in\defEns{0,\ldots,m-1}$, define
\begin{equation*}
  n_{j} = \sum_{i=1}^{\ntest} \1\defEns{\max_{k\in\defEns{1,\ldots,\nbclass}} \pbhat{k}{\sx_i} \in \coint{\alpha_j, \alpha_{j+1}}} \eqsp.
\end{equation*}
For $j\in\defEns{0,\ldots,m-1}$, we define the average accuracy $\Acal_j$ and confidence $\Ccal_j$ in the $j^{\text{th}}$ bin as:
\begin{align*}
  &\Acal_j = \frac{\1\defEns{n_j>0}}{n_{j}} \sum_{i=1}^{\ntest} \1\bigg\{\sy_i = \argmax_{k\in\defEns{1,\ldots,\nbclass}} \pbhat{k}{\sx_i}, \max_{k\in\defEns{1,\ldots,\nbclass}} \pbhat{k}{\sx_i} \in \coint{\alpha_j, \alpha_{j+1}} \bigg\} \eqsp, \\
  &\Ccal_j = \frac{\1\defEns{n_j>0}}{n_{j}} \sum_{i=1}^{\ntest} \max_{k\in\defEns{1,\ldots,\nbclass}} \pbhat{k}{\sx_i} \1\bigg\{\max_{k\in\defEns{1,\ldots,\nbclass}} \pbhat{k}{\sx_i} \in \coint{\alpha_j, \alpha_{j+1}} \bigg\} \eqsp.
\end{align*}
We can then relax equation \eqref{eq:calibration-2} to be: $\Acal_j = \Ccal_j$, for all $j\in\defEns{0,\ldots,m-1}$. When a model does not satisfy $\Acal_j = \Ccal_j$ for all $j$, we say it is miscalibrated. There are a number of ways to measure miscalibration, for example (see \eg~\cite{guo:pleiss:sun:weinberger:2017}):
\begin{itemize}
  \item a reliability diagram, a barplot plotting $\Acal_j$ \wrt~$\Ccal_j$ for every $j\in\defEns{0,\ldots,m-1}$,
  \item the expected calibration error defined as
  \begin{equation*}
    \ECE = \sum_{j=0}^{m-1} \frac{n_{j}}{\ntest} \absolute{\Acal_j - \Ccal_j} \eqsp,
  \end{equation*}
  \item the maximum calibration error defined as
  \begin{equation*}
    \MCE = \max_{j\in\defEns{0,\ldots,m-1}} \absolute{\Acal_j - \Ccal_j} \eqsp.
  \end{equation*}
\end{itemize}

\begin{figure}[tb]
\begin{center}
\includegraphics[scale=0.5]{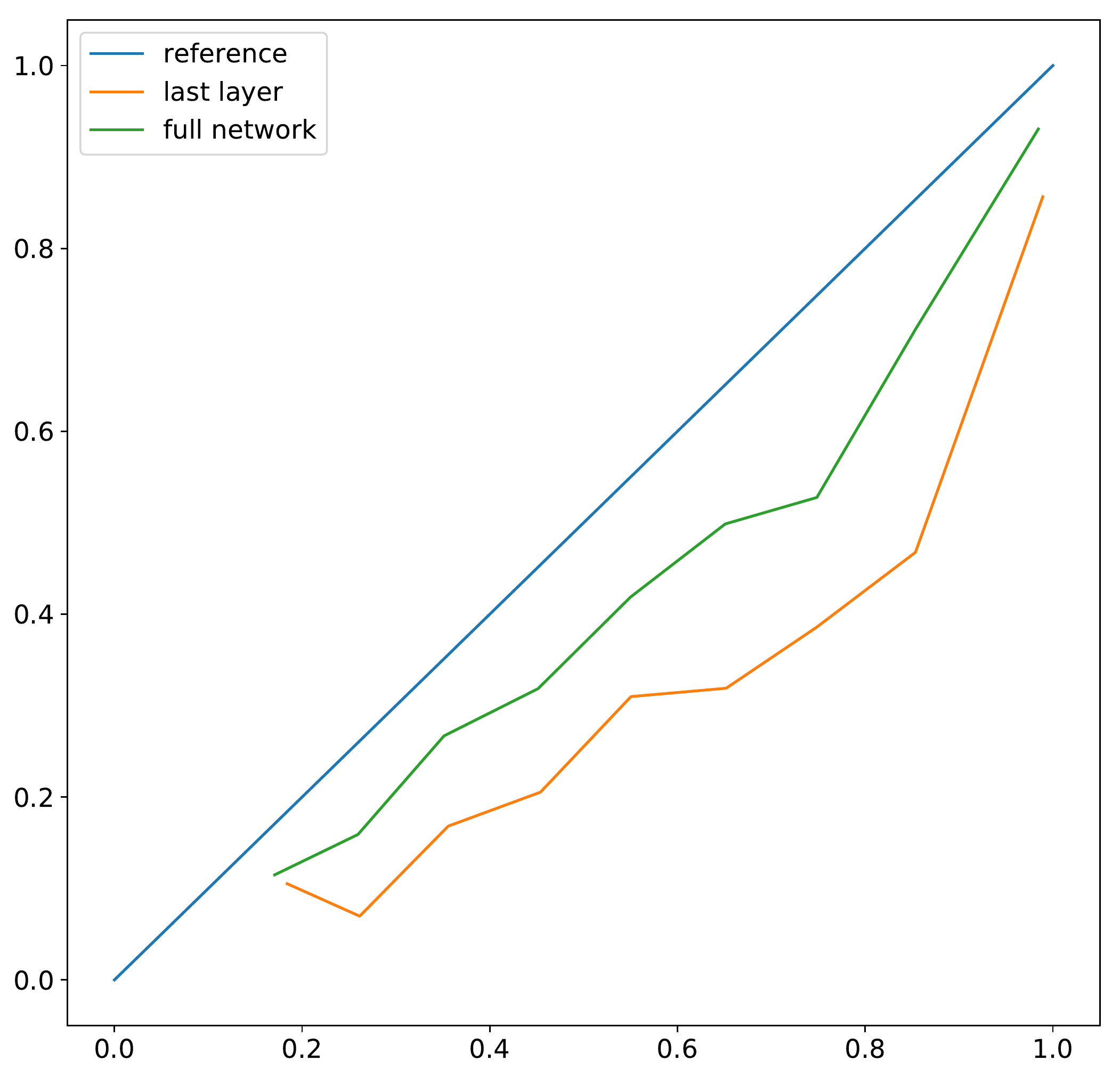}
\caption{
Reliability diagrams for SGD on CIFAR-100. The full network version is better calibrated than the last layer version.
horizontal axis: mean $\max_{k\in\defEns{1,\ldots,\nbclass}} \pbhat{k}{\sx}$ in each bin. vertical axis: empirical accuracy on the test set restricted to each bin.
}
\label{figure:cifar100_pmax_calibration_sgd}
\end{center}
\end{figure}

The reliability diagram using the last layer or full network version of SGD on CIFAR-100 is plotted in \Cref{figure:cifar100_pmax_calibration_sgd}. The full network version of SGD is better calibrated than the last layer version, which is corroborated by an ECE of 0.096 against 0.18. It is consistent with \Cref{figure:cifar100_pmax_hist_sgd} where the full network version shows more dispersed values of $\max_{k\in\defEns{1,\ldots,\nbclass}} \pbhat{k}{\sx}$.

\begin{figure}[ht]
\begin{center}
\includegraphics[scale=0.5]{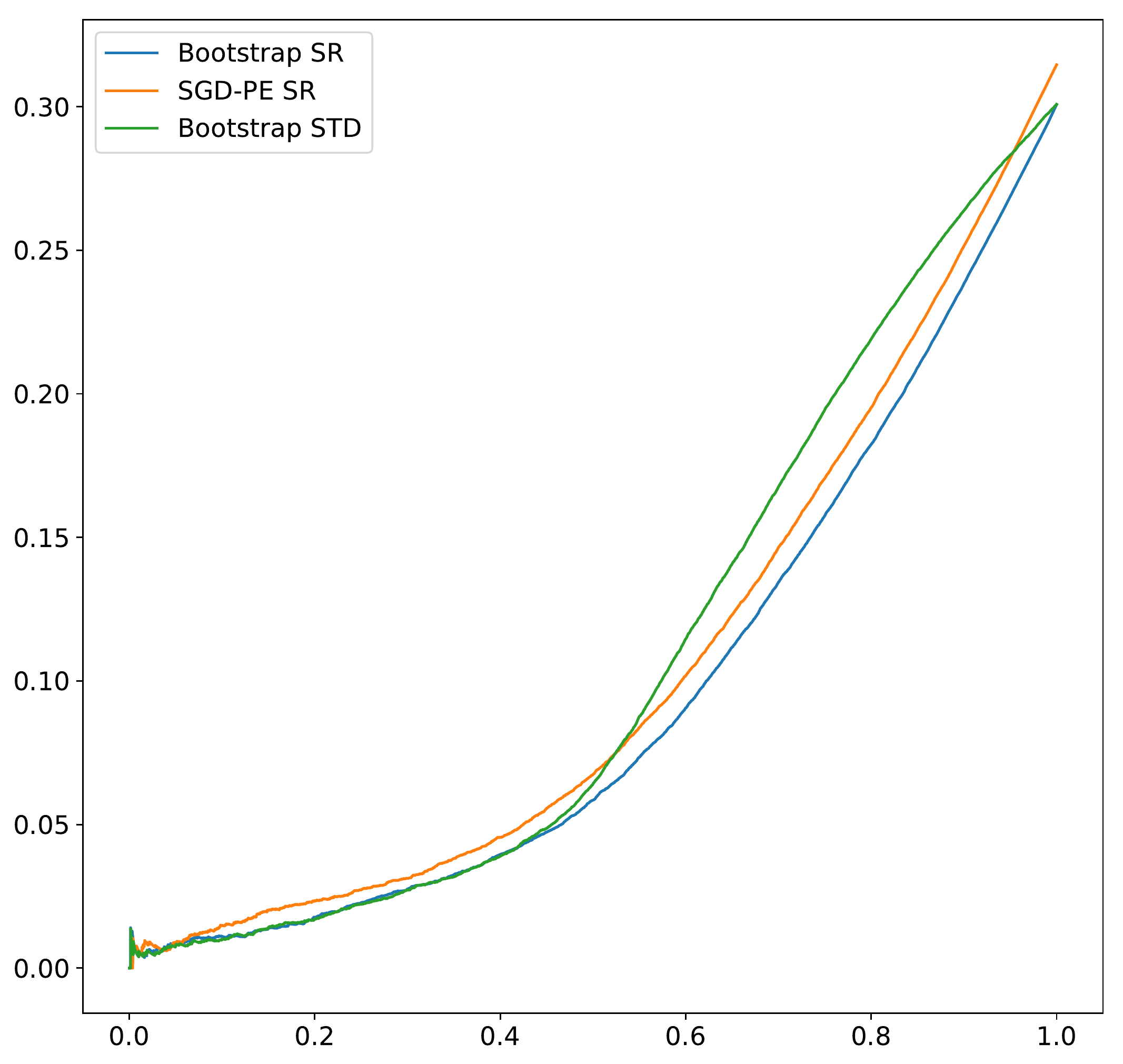}
\caption{
Risk-Coverage curve on ImageNet, for Bootstrap and SGD-PE with two dense layers on top of NASNet. For Bootstrap, 2 confidence functions are compared, $\SRhat$ and $\STDhat$, and for SGD-PE, $\SRhat$.
}
\label{figure:imagenet_risk_coverage}
\end{center}
\end{figure}

In \Cref{figure:imagenet_risk_coverage}, the risk coverage curve on Imagenet of SGD-PE and Bootstrap is displayed. We observe that the confidence function $\SRhat$ of Bootstrap (using the maximum of the predictive posterior distribution as a confidence estimate) achieves a better selection than the confidence function $\SRhat$ of SGD-PE. Besides, the curve also supports the fact that $\SRhat$ is preferable to the $\STDhat$ confidence function.

\begin{figure}[ht]
\begin{center}
\includegraphics[scale=0.5]{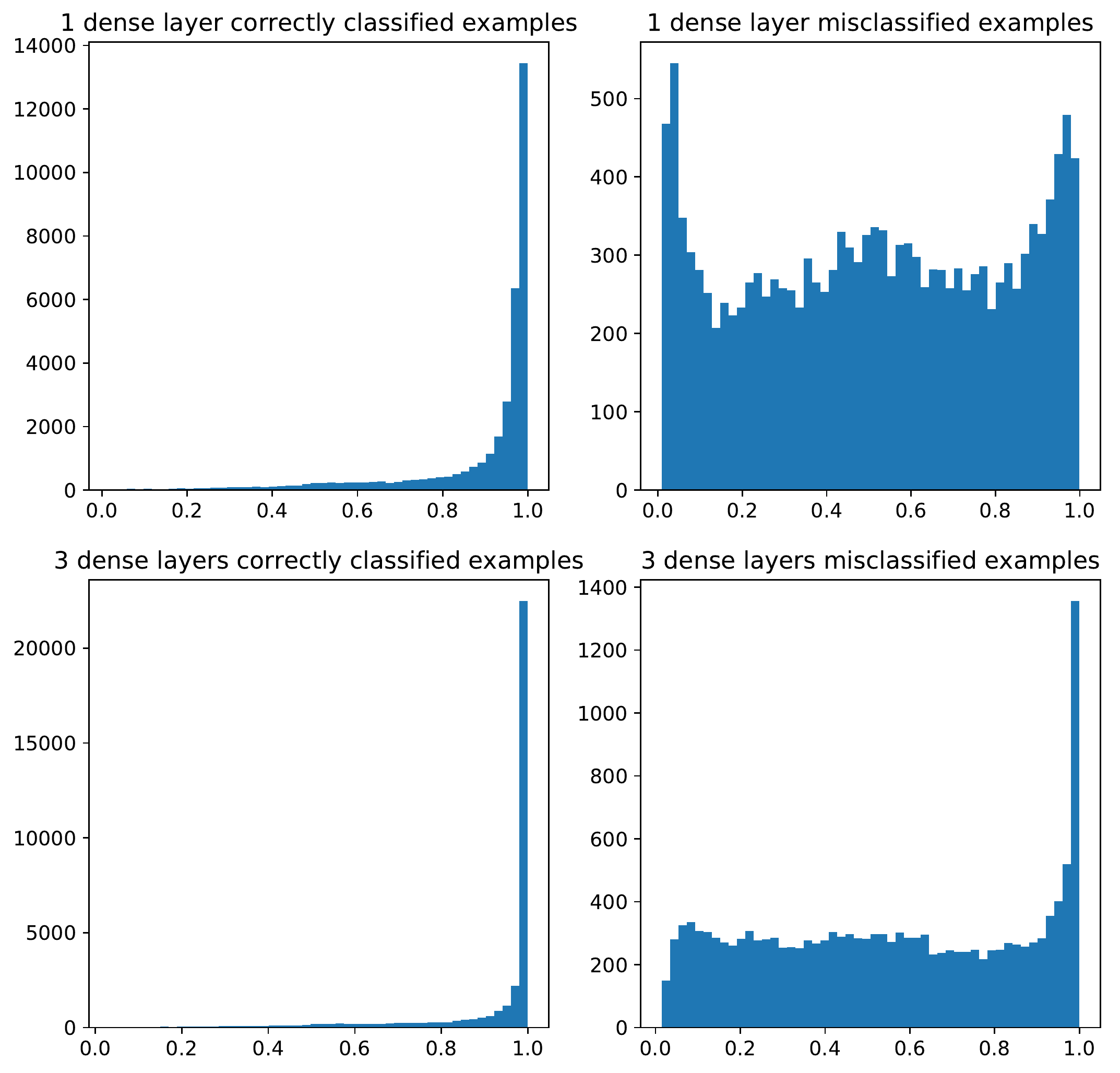}
\caption{
Histograms of the values of the $\SRhat$ confidence function defined in \eqref{eq:def-SR-empirical} for the correctly classified and misclassified test examples.
horizontal axis: $\SRhat$ values. vertical axis: number of items in each bin.
\textbf{Top row:} 1 dense layer on top of NASNet, Bootstrap on ImageNet. Left: histograms for the correctly classified examples. Right: for the misclassified examples.
\textbf{Bottom row:} 3 dense layers on top of NASNet, Bootstrap on ImageNet.
}
\label{figure:imagenet_pmax_hist_bootstrap}
\end{center}
\end{figure}

\begin{figure}[ht]
\begin{center}
\includegraphics[scale=0.5]{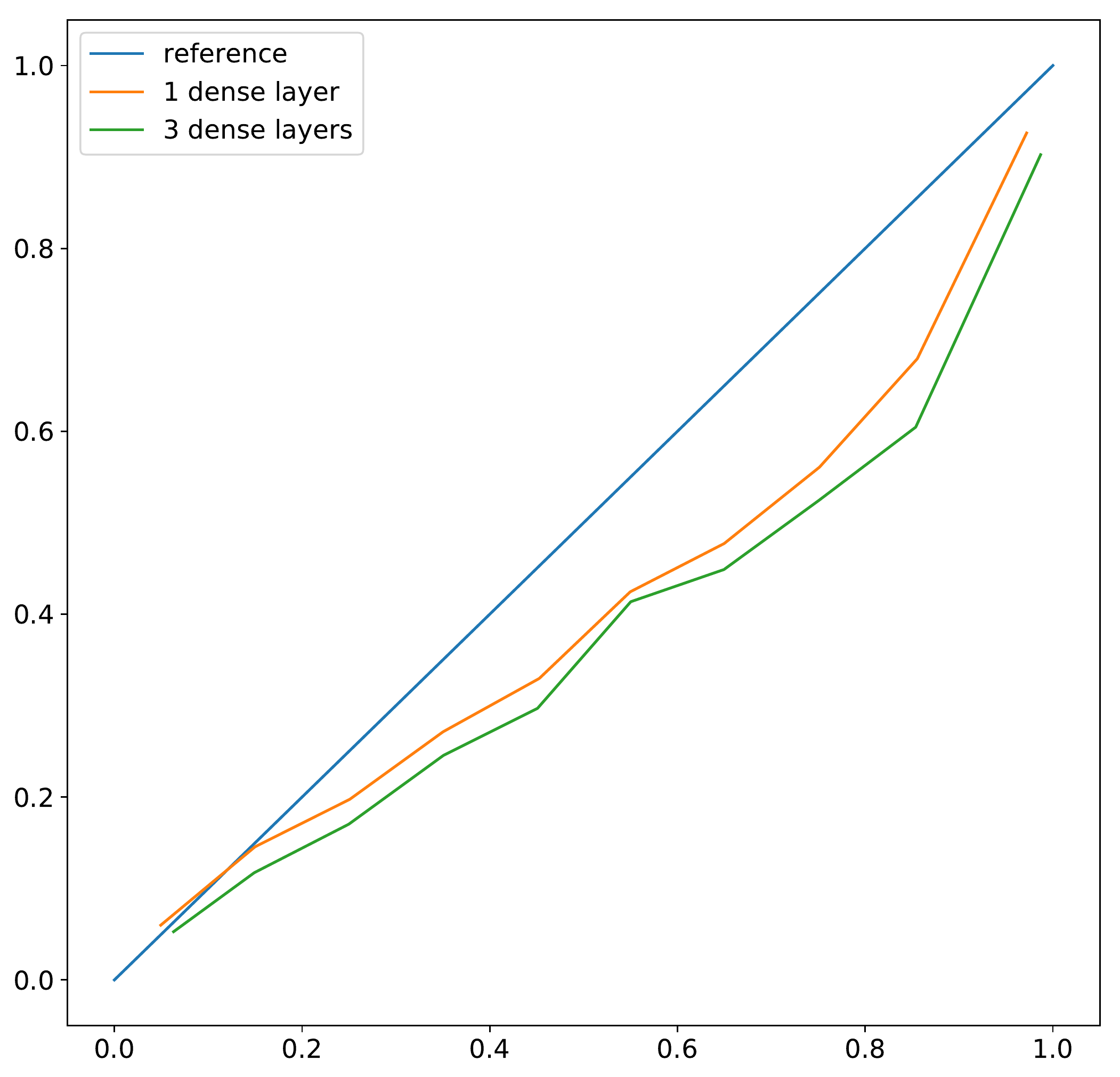}
\caption{
Reliability diagrams for Bootstrap on ImageNet. 1 and 3 dense layer(s) on top of NASNet.
horizontal axis: mean $\max_{k\in\defEns{1,\ldots,\nbclass}} \pb{k}{\sx}$ in each bin. vertical axis: empirical accuracy on the test set restricted to each bin.
}
\label{figure:imagenet_pmax_calibration_bootstrap}
\end{center}
\end{figure}

Following the setup of \Cref{figure:cifar100_pmax_hist_sgd,figure:mnist_pmax_hist_sgd},
\Cref{figure:imagenet_pmax_hist_bootstrap} presents the histograms of the values of the $\SRhat$ confidence function on ImageNet, with 1 or 3 dense layers on top of NASNet, using the Bootstrap algorithm. We observe the inverse phenomenon compared to  \Cref{figure:cifar100_pmax_hist_sgd}: when the number of dense layers at the top of NASNet increases, the histograms of the $\SRhat$ values become \emph{more} concentrated for both the correctly classified and misclassified examples.
In a consistent manner, the 1 dense layer version of Bootstrap is slightly better calibrated than the 3 dense layers version: the reliability diagram is plotted in \Cref{figure:imagenet_pmax_calibration_bootstrap} and the ECE values are 0.08 against 0.11.
This phenomenon may recall some empirical observations \cite{guo:pleiss:sun:weinberger:2017}: when the neural network becomes more complex and grows in size, the model tends to become less calibrated. It emphasizes the importance of the \emph{architecture} of the neural network where the uncertainty algorithms are applied: on a convolutional-type structure, the model seems to become more calibrated; on the inverse, on several stacked dense layers, it tends to lose its calibration property.

\section{Out-of-distribution detection}
\label{sec:suppl-out-of-distribution}

Out-of-distribution detection, \ie~finding out when a data point is not drawn from the training data distribution, is an important and difficult task.
Its importance stems from the fact that we need \emph{robust} models that acknowledge their own limitations.
Detection is hard as high-dimensional probability distributions are challenging to deal with, and often times require unreasonable amounts of data.
Consequently, a flurry of work has been developed \cite{hendrycks:gimpel:17,2018arXiv181205720H,NIPS2018_7915,NIPS2018_7947,
NIPS2018_7967,hendrycks2018deep,2018arXiv180904729S,liang2018enhancing,2018arXiv180204865D,
nalisnick2018do}; unfortunately, describing all of them is beyond the scope of this paper.

\subsection{AUROC and AUPR in/out}
\label{sec:metrics-ood}

Uncertainty estimates are an opportunity to detect \emph{out-of-distribution} samples; with this in mind, the task is reduced to a binary classification (in/out of distribution) and standard metrics like the Area Under Receiver Operating Characteristic curve (AUROC) and the Area Under the Precision-Recall curve (AUPR) can be used, see for example \cite{hendrycks:gimpel:17,liang:li:srikant:2018}.
The in-distribution samples may be treated as the positive class, and the out-of-distribution samples as the negative class (or vice-versa). This binary classification is based on a score and a threshold such that the scores above the threshold are classified as positive and the ones below as negative. In our case, the score is given by a confidence function $\conf:\spacex\to\rset$ and the out-of-distribution samples are supposed to be the least confident inputs according to $\conf$.

Define the true positive rate by $\TPR = \TP / (\TP + \FN)$, and the false positive rate by $\FPR = \FP / (\FP + \TN)$ where $\TP$ is the number of true positive, $\FN$ the number of false negative, $\FP$ the number of false positive and $\TN$ the number of true negative. The ROC curve plots the true positive rate $\TPR$ with respect to the false positive rate $\FPR$ and the AUROC can be interpreted as the probability that a positive example has a greater score than a negative example. Consequently, a random detector corresponds to a $50\%$ AUROC and a perfect classifier corresponds to a $100\%$ AUROC.

The AUROC is not ideal when the positive class and negative class have greatly differing base rates, and the AUPR adjusts for these different positive and negative base rates \cite{Davis:2006:RPR:1143844.1143874,pmid25738806}.
The PR curve plots the precision $\TP /(\TP + \FP)$ and recall $\TP / (\TP + \FN)$ against each other. A random detector has an AUPR equal to the fraction of positive samples in the dataset and a perfect classifier has an AUPR of $100\%$.
Since the baseline AUPR is equal to the fraction of positive samples, the positive class must be specified; in view of this, the AUPRs are displayed when the in-distribution classes are treated as positive (AUPR in), and vice-versa when the out-of-distribution samples are treated as positive (AUPR out).

\subsection{Experimental Results and Discussion}
\label{sec:suppl-auroc-aupr-results} 

Empirical evaluation of out-of-distribution behavior is hard: there are lots of ways to not match the training distribution, some more radical than others.
We want to test reasonably similar out-of-distribution examples.
Thus, we decided to train our models on the first half of the classes, while treating the other half as out-of-distribution samples.
Also, at this point, we do not include ImageNet in our experiments since the full training of NASNet was too computationally intensive, and leave this as future work.

For CIFAR-10/100, we follow a standard training procedure with a decaying learning rate over 250 epochs\footnote{\url{https://github.com/geifmany/cifar-vgg}}.
For MNIST, we use the default Adam optimizer over 20 epochs\footnote{\url{https://keras.io/optimizers/}}.
The point-estimate weights are used both as a reference (SGD-PE), and as the starting point for the last-layer algorithms.
These algorithms, MC-Dropout, SGD, SGLD and Bootstrap, are then trained on (the encoded) half the classes of MNIST and CIFAR-10/100 datasets.

In \Cref{table:mnist-auroc,table:cifar-10-auroc,table:cifar-100-auroc}, we report the results for AUROC.
For completeness, the AUPR out results are shown in \Cref{table:mnist-aupr-out,table:cifar10-aupr-out,table:cifar100-aupr-out} and the AUPR in results are presented in \Cref{table:mnist-aupr-in,table:cifar10-aupr-in,table:cifar100-aupr-in}.

The confidence functions used for the computation of AUROC and AUPR in/out are \textsc{SR}, \textsc{STD} and \textsc{q}, the entropy of $\hat{q}$. They are not reported for AUROC using the entropy of $\hat{q}$ on the MNIST and CIFAR-10 datasets because this confidence function give consistently lower AUROC values.
\textsc{max} is the maximum of the two or three AUROC/AUPR in/out values, and \textsc{increase} is the ratio of the \textsc{max} over the reference SGD-PE. We recall that the higher is the AUROC or AUPR in/out, the better.
The take-away messages are aligned with those from the previous section: last-layer perform comparably to full-network versions, SR dominates other confidence functions, and SGD-PE is a strong contender.


\begin{table*}[t]
	\caption{AUROC for MC-Dropout, Bootstrap, SGD, SGLD and SGD-PE on half classes of MNIST dataset.}
	\label{table:mnist-auroc}
	\vskip 0.15in
	\begin{center}
		\begin{small}
			\begin{sc}
				\begin{tabular}{l|ccc|c|r}
					\toprule
					algorithm & AUROC sr & AUROC std & max AUROC & increase \\
					\midrule
					dropout         & 0.916          & 0.901      & 0.916      & 1.033          \\
					dropout full   & 0.940          & 0.928      & 0.940      & 1.061          \\
					bootstrap       & 0.872          & 0.885      & 0.885      & 0.998           \\
					bootstrap full & 0.898          & 0.908      & 0.909      & 1.025           \\
					sgd             & 0.886          & 0.895      & 0.895      & 1.009           \\
					sgd full       & 0.933          & 0.936      & 0.936      & 1.056           \\
					sgld            & 0.903          & 0.918      & 0.918      & 1.036           \\
					\textbf{sgld full} & \textbf{0.938} & \textbf{0.941} & \textbf{0.941} & \textbf{1.062}           \\
					\midrule
					sgd-pe        & 0.886          &            & 0.886      & 1.000           \\
					\bottomrule
				\end{tabular}
			\end{sc}
		\end{small}
	\end{center}
	\vskip -0.1in
\end{table*}

\begin{table*}[t]
	\caption{AUROC for MC-Dropout, Bootstrap, SGD, SGLD and SGD-PE on half classes of CIFAR-10 dataset.}
	\label{table:cifar-10-auroc}
	\vskip 0.15in
	\begin{center}
		\begin{small}
			\begin{sc}
				\begin{tabular}{l|ccc|c|r}
					\toprule
					algorithm & AUROC sr & AUROC std & max AUROC & increase \\
					\midrule
					dropout         & 0.791          & 0.793      & 0.793      & 1.005          \\
					\textbf{dropout full} & \textbf{0.795} & \textbf{0.792} & \textbf{0.795} & \textbf{1.007}           \\
					bootstrap       & 0.790          & 0.777      & 0.790      & 1.001           \\
					bootstrap full & 0.789          & 0.794      & 0.794      & 1.006           \\
					sgd             & 0.792          & 0.772      & 0.792      & 1.003           \\
					sgd full       & 0.791          & 0.788      & 0.791      & 1.002          \\
					sgld & 0.789 & 0.794 & 0.794 & 1.006           \\
					sgld full      & 0.790          & 0.786      & 0.790      & 1.000           \\
					\midrule
					sgd-pe        & 0.789          &            & 0.789      & 1.000           \\
					\bottomrule
				\end{tabular}
			\end{sc}
		\end{small}
	\end{center}
	\vskip -0.1in
\end{table*}

\begin{table*}[t]
	\caption{AUROC for MC-Dropout, Bootstrap, SGD, SGLD and SGD-PE on half classes of CIFAR-100 dataset.}
	\label{table:cifar-100-auroc}
	\vskip 0.15in
	\begin{center}
		\begin{small}
			\begin{sc}
				\begin{tabular}{l|ccc|c|r}
					\toprule
					algorithm & AUROC q & AUROC sr & AUROC std & max AUROC & increase \\
					\midrule
					dropout         & 0.575    & 0.722          & 0.719      & 0.722      & 1.010           \\
					\textbf{dropout full}   & \textbf{0.736}    & \textbf{0.731}          & \textbf{0.658}      & \textbf{0.736}      & \textbf{1.030}           \\
					bootstrap       & 0.499    & 0.717          & 0.697      & 0.717      & 1.003           \\
					bootstrap full & 0.653    & 0.720          & 0.703      & 0.720      & 1.007           \\
					sgd             & 0.546    & 0.726          & 0.707      & 0.726      & 1.015           \\
					sgd full       & 0.694    & 0.719          & 0.704      & 0.719      & 1.006           \\
					sgld            & 0.599    & 0.728          & 0.718      & 0.728      & 1.018           \\
					sgld full      & 0.576    & 0.713          & 0.710      & 0.713      & 0.998           \\
					\midrule
					sgd-pe        &          & 0.715          &            & 0.715      & 1.000           \\
					\bottomrule
				\end{tabular}
			\end{sc}
		\end{small}
	\end{center}
	\vskip -0.1in
\end{table*}


\begin{table*}[t]
\caption{AUPR out for MC-Dropout, Bootstrap, SGD, SGLD and SGD-PE on half classes of MNIST dataset.}
\label{table:mnist-aupr-out}
\vskip 0.15in
\begin{center}
\begin{small}
\begin{sc}
\begin{tabular}{l|ccc|c|r}
\toprule
algorithm       & AUPR out q & AUPR out sr & AUPR out std & max AUPR out & increase \\
					\midrule
bootstrap       & 0.611        & 0.891              & 0.895          & 0.895          & 0.997               \\
bootstrap full & 0.584        & 0.909              & 0.913          & 0.913          & 1.017               \\
dropout         & 0.880        & 0.911              & 0.897          & 0.911          & 1.014               \\
\textbf{dropout full}   & \textbf{0.869}        & \textbf{0.935}              & \textbf{0.925}          & \textbf{0.935}          & \textbf{1.041}               \\
sgd             & 0.619        & 0.903              & 0.905          & 0.905          & 1.008               \\
sgd full       & 0.599        & 0.932              & 0.932          & 0.932          & 1.038               \\
sgld            & 0.765        & 0.914              & 0.921          & 0.921          & 1.026               \\
sgld full      & 0.925        & 0.934              & 0.932          & 0.934          & 1.040 \\
\midrule
sgd-pe        &              & 0.898              &                & 0.898          & 1.000               \\
\bottomrule
\end{tabular}
\end{sc}
\end{small}
\end{center}
\vskip -0.1in
\end{table*}

\begin{table*}[t]
\caption{AUPR out for MC-Dropout, Bootstrap, SGD, SGLD and SGD-PE on half classes of CIFAR-10 dataset.}
\label{table:cifar10-aupr-out}
\vskip 0.15in
\begin{center}
\begin{small}
\begin{sc}
\begin{tabular}{l|ccc|c|r}
\toprule
algorithm       & AUPR out q & AUPR out sr & AUPR out std & max AUPR out & increase \\
					\midrule
bootstrap       & 0.509        & 0.747              & 0.730          & 0.747          & 0.999               \\
\textbf{bootstrap full} & \textbf{0.521}        & \textbf{0.747}              & \textbf{0.757}          & \textbf{0.757}          & \textbf{1.013}               \\
dropout         & 0.566        & 0.748              & 0.751          & 0.751          & 1.005               \\
dropout full   & 0.651        & 0.749              & 0.744          & 0.749          & 1.002               \\
sgd             & 0.533        & 0.752              & 0.730          & 0.752          & 1.006               \\
sgd full       & 0.692        & 0.755              & 0.754          & 0.755          & 1.010               \\
sgld            & 0.512        & 0.747              & 0.754          & 0.754          & 1.008               \\
sgld full      & 0.573        & 0.749              & 0.749          & 0.751          & 1.004               \\
\midrule
sgd-pe        &              & 0.747              &                & 0.747          & 1.000               \\
\bottomrule
\end{tabular}
\end{sc}
\end{small}
\end{center}
\vskip -0.1in
\end{table*}

\begin{table*}[t]
\caption{AUPR out for MC-Dropout, Bootstrap, SGD, SGLD and SGD-PE on half classes of CIFAR-100 dataset.}
\label{table:cifar100-aupr-out}
\vskip 0.15in
\begin{center}
\begin{small}
\begin{sc}
\begin{tabular}{l|ccc|c|r}
\toprule
algorithm       & AUPR out q & AUPR out sr & AUPR out std & max AUPR out & increase \\
					\midrule
bootstrap       & 0.509        & 0.670              & 0.631          & 0.670          & 1.003               \\
bootstrap full & 0.645        & 0.678              & 0.636          & 0.678          & 1.015               \\
dropout         & 0.594        & 0.682              & 0.673          & 0.682          & 1.022               \\
\textbf{dropout full}   & \textbf{0.703}        & \textbf{0.692}              & \textbf{0.588}          & \textbf{0.703}          & \textbf{1.052}               \\
sgd             & 0.570        & 0.687              & 0.658          & 0.687          & 1.028               \\
sgd full       & 0.668        & 0.674              & 0.635          & 0.674          & 1.009               \\
sgld            & 0.612        & 0.685              & 0.667          & 0.685          & 1.025               \\
sgld full      & 0.589        & 0.668              & 0.654          & 0.668          & 1.000               \\
\midrule
sgd-pe        &              & 0.668              &                & 0.668          & 1.000               \\
\bottomrule
\end{tabular}
\end{sc}
\end{small}
\end{center}
\vskip -0.1in
\end{table*}


\begin{table*}[t]
\caption{AUPR in for MC-Dropout, Bootstrap, SGD, SGLD and SGD-PE on half classes of MNIST dataset.}
\label{table:mnist-aupr-in}
\vskip 0.15in
\begin{center}
\begin{small}
\begin{sc}
\begin{tabular}{l|ccc|c|r}
\toprule
algorithm       & AUPR in q & AUPR in sr & AUPR in std & max AUPR in & increase \\
					\midrule
bootstrap       & 0.550       & 0.817             & 0.841         & 0.841         & 1.002              \\
bootstrap full & 0.539       & 0.855             & 0.873         & 0.873         & 1.041              \\
dropout         & 0.817       & 0.914             & 0.899         & 0.914         & 1.090              \\
\textbf{dropout full}   & \textbf{0.765}       & \textbf{0.938}             & \textbf{0.925}         & \textbf{0.938}         & \textbf{1.119}              \\
sgd             & 0.562       & 0.839             & 0.854         & 0.854         & 1.019              \\
sgd full       & 0.549       & 0.913             & 0.925         & 0.925         & 1.103              \\
sgld            & 0.641       & 0.866             & 0.894         & 0.895         & 1.067              \\
sgld full      & 0.881       & 0.924             & 0.935         & 0.935         & 1.114 \\
\midrule
sgd-pe        &             & 0.839             &               & 0.839         & 1.000              \\
\bottomrule
\end{tabular}
\end{sc}
\end{small}
\end{center}
\vskip -0.1in
\end{table*}

\begin{table*}[t]
\caption{AUPR in for MC-Dropout, Bootstrap, SGD, SGLD and SGD-PE on half classes of CIFAR-10 dataset.}
\label{table:cifar10-aupr-in}
\vskip 0.15in
\begin{center}
\begin{small}
\begin{sc}
\begin{tabular}{l|ccc|c|r}
\toprule
algorithm       & AUPR in q & AUPR in sr & AUPR in std & max AUPR in & increase \\
					\midrule
bootstrap       & 0.503       & 0.811             & 0.799         & 0.811         & 1.001              \\
bootstrap full & 0.510       & 0.809             & 0.813         & 0.813         & 1.003              \\
dropout         & 0.521       & 0.813             & 0.814         & 0.814         & 1.005              \\
\textbf{dropout full}   & \textbf{0.570}       & \textbf{0.818}             & \textbf{0.815}         & \textbf{0.818}         & \textbf{1.009}              \\
sgd             & 0.509       & 0.808             & 0.790         & 0.808         & 0.996              \\
sgd full       & 0.606       & 0.809             & 0.805         & 0.809         & 0.998              \\
sgld            & 0.497       & 0.810             & 0.817         & 0.817         & 1.008              \\
sgld full      & 0.518       & 0.810             & 0.802         & 0.810         & 0.999              \\
\midrule
sgd-pe        &             & 0.810             &               & 0.810         & 1.000              \\
\bottomrule
\end{tabular}
\end{sc}
\end{small}
\end{center}
\vskip -0.1in
\end{table*}

\begin{table*}[t]
\caption{AUPR in for MC-Dropout, Bootstrap, SGD, SGLD and SGD-PE on half classes of CIFAR-100 dataset.}
\label{table:cifar100-aupr-in}
\vskip 0.15in
\begin{center}
\begin{small}
\begin{sc}
\begin{tabular}{l|ccc|c|r}
\toprule
algorithm       & AUPR in q & AUPR in sr & AUPR in std & max AUPR in & increase \\
					\midrule
bootstrap       & 0.499       & 0.737             & 0.717         & 0.737         & 1.000              \\
bootstrap full & 0.607       & 0.740             & 0.732         & 0.740         & 1.004              \\
dropout         & 0.542       & 0.734             & 0.732         & 0.734         & 0.996              \\
dropout full   & 0.737       & 0.741             & 0.701         & 0.741         & 1.005              \\
sgd             & 0.524       & 0.738             & 0.708         & 0.738         & 1.002              \\
sgd full       & 0.660       & 0.736             & 0.731         & 0.736         & 0.999              \\
\textbf{sgld}            & \textbf{0.560}       & \textbf{0.750}             & \textbf{0.737}         & \textbf{0.750}         & \textbf{1.017}              \\
sgld full      & 0.544       & 0.735             & 0.732         & 0.735         & 0.997              \\
\midrule
sgd-pe        &             & 0.737             &               & 0.737         & 1.000              \\
\bottomrule
\end{tabular}
\end{sc}
\end{small}
\end{center}
\vskip -0.1in
\end{table*}

\clearpage

\bibliographystyle{alpha}
\bibliography{../bibliography/bibliographie}

\end{document}